\documentclass{article}
\usepackage{ijcai26}

\usepackage{times}
\usepackage{soul}
\usepackage{url}
\usepackage[hidelinks]{hyperref}
\usepackage[utf8]{inputenc}
\usepackage[small]{caption}
\usepackage{graphicx}
\usepackage{amsmath}
\usepackage{amssymb}
\usepackage{amsthm}
\usepackage{booktabs}
\usepackage{arydshln}
\usepackage{algorithm}
\usepackage{algorithmic}
\usepackage[switch]{lineno}

\usepackage[disable]{todonotes}
\usepackage{placeins}

\usepackage{tabularx}
\usepackage{siunitx}
\usepackage{makecell}

\usepackage{comment}

\title{FlexMoRE: A \underline{Flex}ible \underline{M}ixture \underline{o}f \underline{R}ank-heterogeneous \underline{E}xperts\\
for Efficient Federatedly-trained Large Language Models}

\author{
Annemette Brok Pirchert$^1$
\and
Jacob Nielsen$^{1,2}$\and
Mogens Henrik From$^{1,2}$\and\\
Lukas Galke Poech$^1$\And
Peter Schneider-Kamp$^1$\\
\affiliations
$^1$University of Southern Denmark\\
$^2$Ordbogen A/S\\
\emails
ampirchert@gmail.com\\
\{jacn,from,galke,petersk\}@imada.sdu.dk
}

\begin{document}

\maketitle

\begin{abstract}
Recent advances in mixture-of-experts architectures have shown that individual experts models can be trained federatedly, i.e., in isolation from other experts by using a common base model to facilitate coordination. However, we hypothesize that full-sized experts may not be necessary for all domains and that instead low-rank adapters may be sufficient. Here, we introduce FlexMoRE, a \underline{Flex}ible \underline{M}ixture \underline{o}f \underline{R}ank-heterogenous \underline{E}xperts, which may be either full-sized experts or adapters of a suitable rank. We systematically investigate the trade-off between expert rank and downstream task performance by evaluating $6$ experts with ranks $2^0$ to $2^{14}$ resulting in experiments covering 150 mixtures (96 with 2 experts, 54 with 7 experts) that are evaluated across $120$ tasks. For our experiments, we build on FlexOlmo and turn its pre-trained experts into low-rank versions. Our regression analysis from expert rank to downstream task performance reveals that  the best-performing rank is substantially higher for reasoning-heavy benchmarks than for knowledge-heavy benchmarks. These findings on rank sensitivity come with direct implications for memory efficiency: Using optimal ranks, FlexMoRE yields improved downstream task performance (average score $47.18$) compared to the baseline FlexOlmo-style mixture of full-sized experts (average score $45.46$) at less than one third the parameters ($10.75$B for FlexMoRE vs.\@ $33.27$B for FlexOlmo). All code will be made available.
\end{abstract}

\section{Introduction}
Large language models (LLMs) often benefit from access to domain-specific data in specialized settings. In many practical applications, such data is subject to privacy, legal, or proprietary constraints that limit centralized collection or sharing. At the same time, maintaining multiple domain-adapted models can be costly in terms of hardware, storage, and training resources.

Such constraints commonly arise in domains such as healthcare, law, and enterprise systems, where regulations like the American HIPAA and the European GDPR restrict data movement and usage~\cite{xie2025dflmoedecentralizedfederatedlearning,bdcc9060147}. In these settings, centralized training is often cumbersome or directly impossible. This motivates training approaches and model architectures that can incorporate domain-specific expertise without direct data sharing.

However, combining independently trained models is often non-trivial, as many existing architectures assume joint optimization, shared parameters, or fixed model composition.
Existing approaches address only parts of this problem. Mixture-of-Experts (MoE) architectures scale model capacity via sparse routing  but typically rely on centrally trained full-size experts with a large parameter count and high memory requirements. \cite{cao2024moelightninghighthroughputmoeinference,mu2025comprehensivesurveymixtureofexpertsalgorithms,zhao2025puzzlemoeefficientcompressionlarge}. 
Pathway Language Models (PaLM) \cite{chowdhery2023palm,anil2023palm} orchestrate a model across many accelerators, with models that can generalize over different domains and tasks while being highly efficient. Pathways enable data parallelism at pod (node) level, making it possible to orchestrate data in separate training nodes.
Parameter-efficient fine-tuning methods such as LoRA~\cite{hu2022lora} and Mixture-of-Adapters (MoA)~\cite{cao2025moaheterogeneousmixtureadapters,wang2022adamix} reduce adaptation cost by introducing low-rank, additive modules into a shared backbone. These approaches do not support independently trained experts or inference-time opt-in and opt-out \cite{hu2023llmadaptersadapterfamilyparameterefficient}, though. 
Recently, FlexOlmo~\cite{shi2025flexolmoopenlanguagemodels} enabled decentralized expert training and inference-time composition without data sharing by training domain-specific experts alongside a public and frozen base model. However, a major drawback is that FlexOlmo relies on full-size experts, which in practice limits scalability due to high accelerator memory requirements.

In this work, we introduce FlexMoRE, which builds on the FlexOlmo framework for decentralized expert composition but significantly reduces the parameter count and accelerator memory footprint of individual experts. FlexMoRE supports independently trained full-size and low-rank experts within the same MoE routing framework. The low-rank experts might be either trained as adapters from scratch alongside the public base model, or, they can be derived via post-hoc low-rank factorization of fully-finetuned experts relative to a base expert. The latter we designate as \emph{Post-hoc Low-Rank Adaptation (PHLoRA) experts} \cite{vasani2025phloradatafreeposthoclowrank}.
In our FlexMoRE architecture, low-rank experts are implemented as independent MoE experts relative to a full-size base expert. In the case of FlexOlmo, we can conviently use the public base model as the full-size base expert. In this paper, we show that decentralized training and flexible inference-time composition extend to experts parameterized via low-rank approximations rather than full-size experts.

In sum, our contributions are:
\begin{itemize}
    \item A flexible mixture of rank-heterogeneous experts architecture, called FlexMoRE, in which full-size experts can be deliberately combined with low-rank experts.\footnote{Code will be made available for reproducibility and reuse.}
    \item Empirical support confirming the effectiveness of this architecture obtained through deriving post-hoc LoRA experts from the existing FlexOlmo model, showing without the need for training that we can retain and even improve the performance across most benchmarks at one third of the memory requirements.
    \item A comprehensive study of using different LoRA ranks for the experts and an in-depth analysis of rank sensitivity showing that reasoning-heavy tasks require higher ranks than knowledge-heavy tasks.
\end{itemize}

\begin{figure}[t]
    \centering
    \includegraphics[width=0.6\columnwidth]{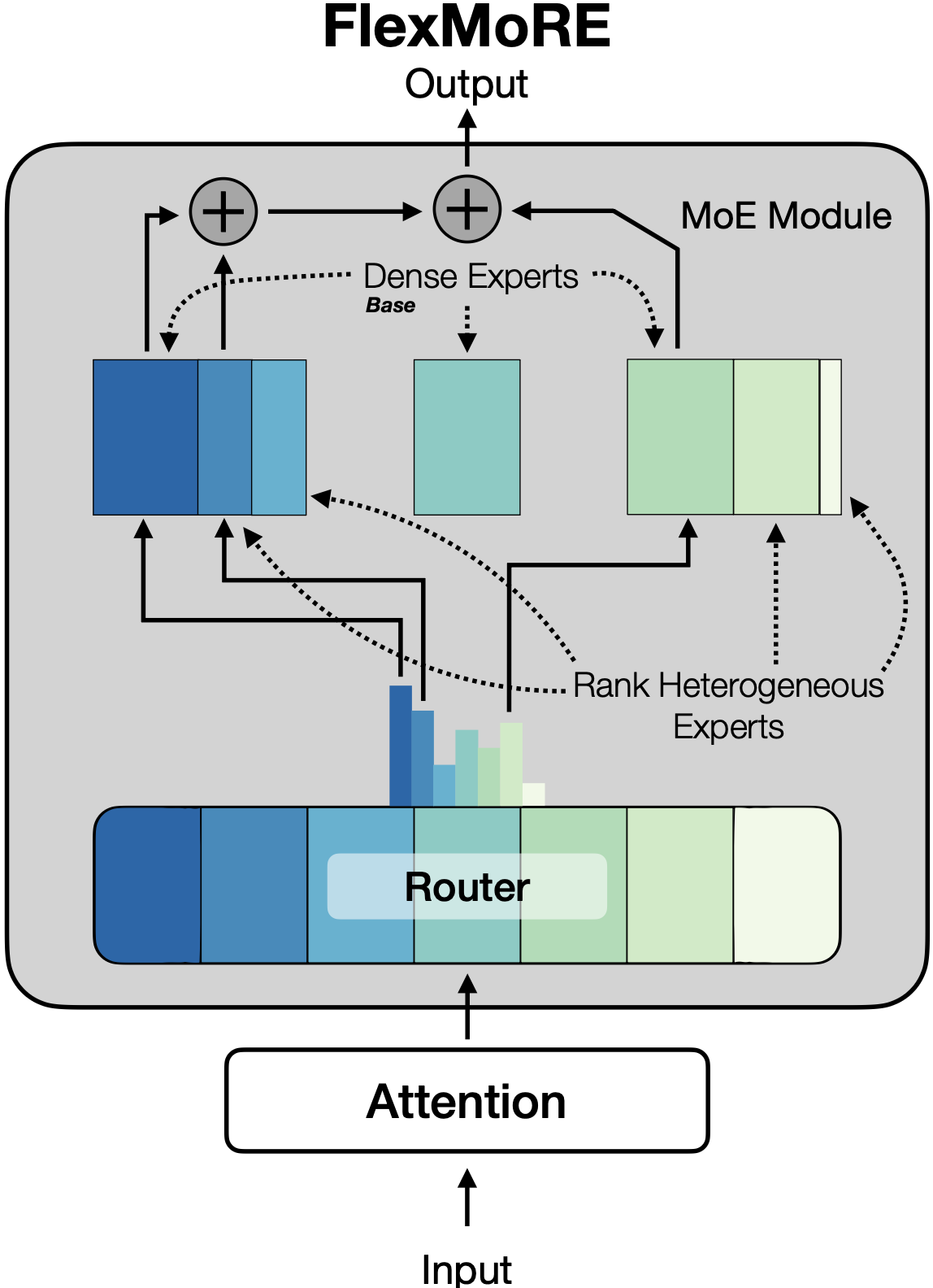}
    \caption{FlexMoRE follows a standard MoE architecture, similarity FlexOlmo, utilizing the domain-informed router, but routing to one or more group(s) with base expert and rank-heterogeneous experts.}
    \label{fig:methods:flexmore}
\end{figure}

\section{Related Work}

\paragraph{Mixture of Experts (MoE)}
MoE architectures enable conditional computation by sparsely routing inputs to a subset of specialized experts, allowing model capacity to scale without proportional increases in per-token computation \cite{fedus2022switch,mu2025comprehensivesurveymixtureofexpertsalgorithms}. MoE architectures are widely used to improve efficiency and specialization in large language models. A central challenge in MoE systems is expert routing, as suboptimal routing can lead to under-utilized or over-specialized experts. Prior works have addressed this through improved routing strategies that aim to stabilize training and improve expert utilization \cite{liu2024deepseek,zhou2022mixtureofexpertsexpertchoicerouting}. Other studies further show that MoE architectures can be made parameter-efficient by restricting updates to lightweight experts, achieving performance comparable to full fine-tuning while modifying only a small fraction of parameters~\cite{zadouri2023pushingmixtureexpertslimit}. System-level optimizations reduce the runtime cost of executing MoE models through memory and inference improvements \cite{rajbhandari2022deepspeedmoeadvancingmixtureofexpertsinference,cao2024moelightninghighthroughputmoeinference,Wang_2025,zhao2025puzzlemoeefficientcompressionlarge} but assume a fixed, centrally trained pool of experts and shared data access.

\paragraph{Low-Rank Adapters and Mixtures}
Parameter-efficient fine-tuning methods adapt large language models by introducing lightweight trainable modules while keeping the backbone frozen. Low-Rank Adaptation (LoRA)~\cite{hu2022lora} is a widely adopted approach that substantially reduces training cost and the number of trainable parameters while maintaining strong performance. Multiple LoRA adapters can be treated as expert models that can selectively be routed to \cite{hu2023llmadaptersadapterfamilyparameterefficient}. Several recent works extend LoRA using MoE designs, exploring different routing and expert allocation strategies such as dynamic routing, layer-wise expert placement, and sparse expert activation \cite{kunwar2025ttloramoeunifyingparameterefficient,cao2025moaheterogeneousmixtureadapters,ji2025lmoeendtoendtraininglightweight,zhuang2025ldmolelearnabledynamicrouting,li2024mixloraenhancinglargelanguage,zou2025flyloraboostingtaskdecoupling}. Despite these advances, most MoE–LoRA approaches assume centralized training and joint optimization over a shared adapter pool, limiting their applicability in settings with restricted data sharing and federatedly trained experts. 

\paragraph{Decentralized Expert Composition under Data Governance Constraints}
FlexOlmo introduces a language model architecture designed to support flexible data usage under strict governance constraints, commonly arising from regulations such as HIPAA, GDPR, as well as data sovereignty requirements \cite{shi2025flexolmoopenlanguagemodels,bdcc9060147}. In many real-world settings, these constraints effectively preclude centralized access to sensitive data. As a result, decentralized training paradigms such as federated learning have gained attention, as they avoid centralized access to raw data \cite{Abishek_2025}. MoE variants have also been explored in federated contexts as a means of balancing data heterogeneity, privacy, and model performance \cite{yi2024pfedmoedatalevelpersonalizationmixture}.
Unlike most conventional MoE architectures, which are typically trained via joint optimization over shared or centrally accessible datasets, FlexOlmo enables experts to be trained independently on closed or private data and composed only at inference time. A domain-informed routing mechanism allows experts to be selectively included or excluded during inference, enabling fine-grained control over which data sources contribute to a given prediction.

\paragraph{Summary}
Existing work on MoE, parameter-efficient fine-tuning, and decentralized training addresses orthogonal aspects of model scaling and adaptation.
LoRA-based methods reduce adaptation cost within shared backbones, while FlexOlmo enables inference-time composition of independently trained dense experts under data governance constraints.
With FlexMoRE, we now explore whether FlexOlmo-style decentralized expert composition remains effective with low-rank rather than full-size experts.

\section{Methods}

Here, we present FlexMoRE, our proposed approach for rank-heterogenous Mixture-of-Experts for federated learning of language models. Figure~\ref{fig:methods:flexmore} shows an overview of the architecture.

\subsection{The FlexMoRE architecture}
In a standard MoE architecture, the feedforward module (FFN) in each transformer block is replaced by a routing module and $n$ FFN experts.
At inference time, a limited selection of experts is selected per layer to facilitate a forward pass. 
Experts are typically trained jointly in MoE models.
FlexOlmo~\cite{shi2025flexolmoopenlanguagemodels} introduced the possibility of training experts independently from each other: Each expert is trained only with the globally shared public expert. This ensures that the individual routing modules are anchored against the same base model.

In our proposed FlexMoRE architecture, we additionally consider low-rank experts. We allow that low-rank adapters can be integrated as experts and that the low-rank adapters can be attached to an arbitrary full-size expert. In a FlexOlmo setting, a natural choice for this full-size expert is the public model. This is because the public expert already needs to be consulted for individual expert training. Therefore, the public expert also forms an ideal base for the low-rank adapters.

In theory, FlexMoRE could be composed of an arbitrary mix of at least one full-size and multiple low-rank experts. Each low-rank expert needs to track which other expert (either full-size, or low-rank) it uses as a base. To ensure this relationship is well-founded, we require that at some point, a full-size expert is reached.

Crucially, the low-rank experts do not need to have the same rank. Given the full-size expert $M_\text{base}$, it can have $i$ low-rank adapters, each with a different rank $r_i$, that attach to it $m_i^{r_i}$.
We denote such a combination as $\mathcal{M} := \left( M_\text{base}, \{ m_i^{r_i}\}\right)$, where $m_i$ denotes an expert with index $i$ and rank $r_i$.
One can see that one could compose multiple combinations of such full-rank plus low-rank experts $\mathcal{M}_0, \ldots \mathcal{M}_j$. For the sake of simplicity, however, in this paper, we will only consider cases with a single full-sized expert and sets of low-rank experts $i$ of (potentially varying) ranks $r_i$.

Just as in FlexOlmo, FlexMoRE depends on the domain-informed router function $f$ mapping the input vector $x$ to a distribution over expert modules: $f(x)=W_rx$, $W_r \in \mathbb{R}^{(n+1) \times h}$. This also means we inherit the decomposing of $W_r$ into individual expert-specific router embeddings, having each row representing a specific  full-size $M$ or low-rank $m_i^{r_i}$ expert. Routing to any low‑rank expert also triggers its corresponding full-size base expert, to which we then apply the low-rank adapter $m_i^{r_i}$.

\subsection{Deriving Experts through Adapter Extraction}\label{sec:method:low_rank_experts}
We consider the weights of a public base expert (layer indices omitted for simplicity): $\text{Expert 0 (public)}:  \mathbf{W}_\mathrm{base} \in \mathbb{R}^{d_{\text{out}}\times d_{\text{in}}}$
and the respective expert $i <N$: $\text{Expert } i \text{ (domain)}: \mathbf{W}_i \in \mathbb{R}^{d_{\text{out}}\times d_{\text{in}}}$.
Each trained expert requires identical amounts of memory and computational resources, even when data might not necessitate the entire offered capacity. We hypothesize that some domains require less capacity than others. 

More specifiaclly, we expect that a low rank approximation of an expert $E_{n}$ trained on a domain-specific dataset $D_{\text{domain}}$ is sufficient to perform comparably to a full-size expert. We integrate such experts as follows.
First, we calculate the difference $\boldsymbol{\Delta}_n$ between the public model $\mathbf{W}_0$ and the domain expert $\mathbf{W}_i$: 
$\boldsymbol{\Delta}_i = \mathbf{W}_i - \mathbf{W}_0$.
This represents the contribution of the domain-specific expert, which also implies that each expert remains anchored to their corresponding base expert. Next, we compute the truncated singular value decomposition:
$\boldsymbol{\Delta}_i = \mathbf{U}\boldsymbol{\Sigma}\mathbf{V}^{\top}$,
where $\mathbf{U} \in \mathbb{R}^{d_{\text{out}}\times d_{\text{out}}}$, $\boldsymbol{\Sigma} \in \mathbb{R}^{d_{\text{out}}\times d_{\text{in}}}$, and $\mathbf{V}^{\top} \in \mathbb{R}^{d_{\text{in}}\times d_{\text{in}}}$, denoting the left-singular vector, singular value matrix and left-singular vector, respectively.
We obtain the low-rank approximation by utilizing only the first $r$ values, truncating the SVD, $\boldsymbol{\Delta}_n$, obtaining $\mathbf{U}_r, \boldsymbol{\Sigma}_r$ and $\mathbf{V}_r^{\top}$.
\begin{align}
\mathbf{U}_r &= \mathbf{U}[:, :r] \in \mathbb{R}^{d_{\text{out}}\times r} \\[2pt]
\boldsymbol{\Sigma}_r &= \text{diag}(\sigma_1, \ldots, \sigma_r) \in \mathbb{R}^{r\times r} \\[2pt]
\mathbf{V}_r^{\top} &= \mathbf{V}^{\top}[:r, :] \in \mathbb{R}^{r\times d_{\text{in}}}
\end{align}
This exactly reduces the shape of our expert, defined by the the best rank $r$ approximation. The original matrix can be reconstructed as follows:
\begin{equation}
\widetilde{\boldsymbol{\Delta}}_i^{(r)} = \mathbf{U}_r \boldsymbol{\Sigma}_r \mathbf{V}_r^{\top} = (\mathbf{U}_r \cdot \boldsymbol{\Sigma}_r) \mathbf{V}_r^{\top}
\end{equation}
Thies yields a rank $r$ approximation of the original $\boldsymbol{\Delta}_n$ matrix. 
We employ our rank-tuned expert by adding it to it's corresponding base expert (e.g. the public base model):
\begin{equation}\label{sec:method:eq:base_appx_r}
\widetilde{\mathbf{W}}_i^{(r)} = \mathbf{W}_0 + \widetilde{\boldsymbol{\Delta}}_i^{(r)}
\end{equation}
With Equation \ref{sec:method:eq:base_appx_r}, we formulate an approximation of a given Expert $n$ with an approximation error introduced by truncating the decomposition at rank $r$.

Our architectural design, therefore, allows us to derive low-rank experts from the full-sized experts via PHLoRA~\cite{vasani2025phloradatafreeposthoclowrank}. Here, we obtain SVD components $A$ and $B$ that we need to derive a low-rank expert by splitting the $\Sigma_r$ symmetrically between the $U_r$ and $V_r$ components:
\begin{equation}
    B = U_r \sqrt{\Sigma_r}, A= V_r \sqrt{\Sigma_r}
\end{equation}
This allows us to easily integrate with existing LoRA libraries such as the PEFT Library. This implies the MoE module can compute the output $y$ given a an input $x$ as follows:
\begin{equation}
    y = \Sigma_{i \in \text{Top}k(f(x))} \text{softmax}(f(x)_i)(M_n+B_nA_n)
\end{equation}
where $B$ and $A$ is the components from low rank expert $m \in M$ and $f$ denotes the router function that computes the probabilities from $x$. 
Low-rank adapters could also be trained from scratch in a fashion similar to the full-finetuning performed in FlexOlmo.

\subsection{Rank Sensitivity Analysis}\label{sub:rank-sensitivity-analysis}
A crucial objective of this research is to  understand the relationship between expert rank and task performance.
To quantify this, we estimate rank sensitivity using linear regression between $\log_2$ of the expert rank and task performance.
For each model family and evaluation group, we fit
\label{eq:rank_regression}
$s(r) = \alpha + \beta \log_2 r$,
where $s(r)$ denotes the evaluation score at rank $r$ while $\beta$ measures sensitivity to rank increases.
Positive values of $\beta$ indicate that increasing rank consistently improves performance, while values near zero or negative indicate diminishing returns.
We apply this procedure to both the full combined mixture-of-experts model and the individual experts. 
\label{eq:peak_rank}
To identify the rank at which performance peaks, we further define the \textbf{typical peak rank} for expert $e$ and evaluation group $g$ as
$r^*_{e,g}
= \operatorname*{arg\,max}_{r \in \{2^0,\dots,2^{14}\}} s_{e,g}(r),
$
computed directly from the observed scores without regression, smoothing, or normalization.
In the case of ties, the lowest rank achieving the maximum score is selected.

\section{Experimental Setup}
The purpose of our experiments is to show that the FlexMoRE architecture with low-rank experts does not degrade the performance when compared to full-sized experts.

\subsection{Evaluation Datasets}
We evaluate all models on a benchmark suite aligned with that used in FlexOlmo \cite{shi2025flexolmoopenlanguagemodels}, enabling comparability. Our evaluation spans a diverse collection of established benchmarks grouped into general-purpose and domain-specific evaluations, covering a total of 120 tasks.

\textbf{General-purpose evaluation} includes:
\textbf{MC9}, a collection of nine multiple-choice reasoning benchmarks
(ARC-Easy, ARC-Challenge \cite{clark2018thinksolvedquestionanswering}, BoolQ \cite{clark2019boolqexploringsurprisingdifficulty}, CommonsenseQA (CSQA) \cite{reddy2019coqaconversationalquestionanswering}, HellaSwag \cite{zellers2019hellaswagmachinereallyfinish}, OpenBookQA \cite{mihaylov2018suitarmorconductelectricity}, PIQA \cite{bisk2019piqareasoningphysicalcommonsense}, SocialIQA \cite{sap2019socialiqacommonsensereasoningsocial}, and WinoGrande \cite{sakaguchi2019winograndeadversarialwinogradschema});
\textbf{GEN5}, consisting of five generative question answering tasks
(CoQA) \cite{reddy2019coqaconversationalquestionanswering}, SQuAD \cite{rajpurkar2016squad100000questionsmachine}, Natural Questions \cite{kwiatkowski-etal-2019-natural}, TriviaQA \cite{joshi2017triviaqalargescaledistantly}, and DROP \cite{dua2019dropreadingcomprehensionbenchmark});
\textbf{AGIEval}, a suite of college-level academic reasoning tasks
\cite{zhong2023agievalhumancentricbenchmarkevaluating};
and \textbf{BBH}, a collection of challenging multi-step reasoning tasks from BIG-Bench \cite{suzgun2022challengingbigbenchtaskschainofthought}.

\textbf{Domain-specific evaluation} includes
\textbf{MMLU} \cite{hendrycks2021measuringmassivemultitasklanguage}
and \textbf{MMLU-Pro} \cite{Wang2024MMLUProAM},
which both assess broad academic knowledge across multiple disciplines.
In total, we consider 120 individual evaluation tasks.

\subsection{Evaluation Procedure and Baselines}
\label{sec:evaluation-measures-baselines}

\paragraph{Measures} We summarize downstream task performance by an aggregate score `\textit{Avg}', which
follows the same evaluation protocol as FlexOlmo. Specifically, \textit{Avg} is
computed as the unweighted mean over evaluation group means:
\label{eq:avg_score}
$\textit{Avg}
= \frac{1}{|\mathcal{G}|}
\sum_{g \in \mathcal{G}}
\left(
\frac{1}{|\mathcal{T}_g|}
\sum_{t \in \mathcal{T}_g}
s_t
\right),$
where
$\mathcal{G}$ is the set of our six considered evaluation groups (MC9, GEN5, AGIEval, BBH, MMLU, and MMLU-Pro), $\mathcal{T}_g$ is the set of tasks within group
$g$, and $s_t$ is the task-level score. 

\paragraph{Tested configurations}
We evaluate the following configurations: (iii) \textbf{The individual low-rank experts} derived from the six available
FlexOlmo experts (Code, Creative Writing, Math, News, Academic, Reddit), which we evaluate as mixture-of-experts models with two experts. The Educational Text expert is not publicly available and has, thus, not been part of our experimental setup.
(ii) \textbf{homogeneous FlexMoRE} models, with one full-sized expert (the public model) and the remaining experts as low-rank adapters of identical rank. We evaluate ranks from $2^0$ to $2^{11}$, as ranks greater than $2^{11}$ increase rather than decrease the total number of model parameters.
(iii) \textbf{heterogeneous FlexMoRE}, where the low-rank adapters can be of different ranks. To determine the best-performing ranks per expert, we consider their performance on the MC9 eval group or the average \textit{Avg} over all six groups. For MC9 as the reference eval group, we obtain rank $2^6$ for Code, $2^7$ for Creative Writing, $2^{11}$ for Math, $2^0$ for News, $2^3$ for Academic, and $2^7$ for Reddit. When we consider performance on all eval groups, we obtain rank $2^9$ for Code, $2^4$ for Creative Writing, $2^{11}$ for Math and Academic, $2^6$ for News, and $2^9$ for Reddit.

Note that this evaluation procedure differs from the one in FlexOlmo~\cite{shi2025flexolmoopenlanguagemodels}:  All our expert evaluations are conducted in a 2x7B setup with the expert alongside the public base experts.  FlexOlmo instead evaluates single experts in isolation without the public base expert.
Our approach ensures that the evaluation corresponds to both how the experts have been trained and how they are going to be used. For comparability to the full-size experts, we also re-evaluate the six available FlexOlmo experts in the exact same way.

\paragraph{Baselines}
Our main baseline for the final MoE model is FlexOlmo with full-size experts, under varying numbers of active experts (2, 4, 7), reflecting choices reported in \cite{shi2025flexolmoopenlanguagemodels}.
Our baselines for the evaluation of individual low-rank experts are the corresponding experts in their full-size version.
\textbf{Relative improvement} against the respective baseline is quantified as: $\Delta[\%] = 100 \times
\frac{\textit{Avg}_{\textit{model}} - \textit{Avg}_{\text{baseline}}}
     {\textit{Avg}_{\text{baseline}}}$.

\section{Results}
We first present our results with single low-rank experts in Section~\ref{sec:result:comparing_experts}. Then, we report the results for our FlexMoRE models equipped with rank-homogeneous and rank-heterogeneous experts in Section \ref{sec:downstream-baseline-analysis}.

\begin{table*}[h!]
\centering
\small
\sisetup{table-format=1.4}
\begin{tabular*}{\textwidth}{
@{\extracolsep{\fill}}
l
rr|
*{6}{S}
|S
r
}
\toprule
\textbf{2x7B-1T Expert} & {\textbf{Total}} & {\textbf{Expert}} &
{\textbf{MC9}} & {\textbf{GEN5}} & {\textbf{AGIEval}} & {\textbf{BBH}} &
{\textbf{MMLU}} & {\textbf{MMLU-Pro}} &
{\textbf{Avg}} & {\textbf{$\Delta$(\%)}} \\
\midrule
Code (baseline) & 11.63B & 4.33B 
& 0.6757 & 0.4668 & \textbf{0.3909} & \textbf{0.3831} & 0.5381 & 0.2443 & 0.4498 & -- \\\cdashline{1-11}\addlinespace[0.4ex]
Code ($r=2^9$) & \textbf{8.04B} & \textbf{743M} 
& \textbf{0.6866} & \textbf{0.4946} & 0.3824 & 0.3704 & \textbf{0.5464} & \textbf{0.2492} & \textbf{0.4549} & \textbf{+1.14} \\
\cmidrule(lr){1-11}
Creative Writing (baseline) & 11.63B & 4.33B 
& 0.6709 & 0.4998 & \textbf{0.3965} & 0.3424 & 0.5378 & 0.2478 & 0.4492 & -- \\\cdashline{1-11}\addlinespace[0.4ex]
Creative Writing ($r=2^4$) & \textbf{7.32B} & \textbf{23.3M} 
& \textbf{0.6854} & \textbf{0.5074} & 0.3959 & \textbf{0.3545} & \textbf{0.5601} & \textbf{0.2595} & \textbf{0.4605} & \textbf{+2.50} \\
\cmidrule(lr){1-11}
Math (baseline) & 11.63B & 4.33B 
& \textbf{0.6836} & 0.4862 & \textbf{0.4138} & \textbf{0.4589} & \textbf{0.5648} & \textbf{0.2615} & \textbf{0.4781} & -- \\\cdashline{1-11}\addlinespace[0.4ex]
Math ($r=2^{11}$) & \textbf{10.27B} & \textbf{2.97B} 
& 0.6832 & \textbf{0.4954} & 0.4072 & 0.4289 & 0.5557 & 0.2598 & 0.4717 & -1.34 \\
\cmidrule(lr){1-11}
News (baseline) & 11.63B & 4.33B 
& 0.6602 & 0.5058 & 0.3691 & 0.3554 & 0.5387 & 0.2446 & 0.4457 & -- \\\cdashline{1-11}\addlinespace[0.4ex]
News ($r=2^6$) & \textbf{7.39B} & \textbf{92.9M} 
& \textbf{0.6858} & \textbf{0.5078} & \textbf{0.4004} & \textbf{0.3646} & \textbf{0.5578} & \textbf{0.2624} & \textbf{0.4631} & \textbf{+3.92} \\
\cmidrule(lr){1-11}
Academic (baseline) & 11.63B & 4.33B 
& \textbf{0.6732} & 0.5000 & \textbf{0.3944} & 0.3622 & 0.5479 & 0.2422 & 0.4533 & -- \\\cdashline{1-11}\addlinespace[0.4ex]
Academic ($r=2^{11}$) & \textbf{10.27B} & \textbf{2.97B} 
& 0.6710 & \textbf{0.5104} & 0.3928 & \textbf{0.3653} & \textbf{0.5486} & \textbf{0.2515} & \textbf{0.4566} & \textbf{+0.73} \\
\cmidrule(lr){1-11}
Reddit (baseline) & 11.63B & 4.33B 
& 0.6171 & 0.4108 & 0.3668 & \textbf{0.3668} & 0.5358 & 0.2350 & 0.4221 & -- \\\cdashline{1-10}\addlinespace[0.4ex]
Reddit ($r=2^9$) & \textbf{8.04B} & \textbf{743M} 
& \textbf{0.6768} & \textbf{0.5018} & \textbf{0.3983} & 0.3499 & \textbf{0.5441} & \textbf{0.2422} & \textbf{0.4522} & \textbf{+7.13} \\
\bottomrule
\end{tabular*}
\caption{
Best post-hoc LoRA experts with rank $r = 2^k,\; k \in \{0,\dots,11\}$
 compared to their full-size baselines. Model size is reported as number of total parameters of the 2x7B mixture and for the expert only.
All scores are reported as mean performance, and relative improvements ($\Delta$ Avg / $\Delta$(\%)) are computed with respect to cl corresponding baseline model.}
\label{sec:results:table:best_expert}
\end{table*}

\begin{comment}
\emph{News} and \emph{Educational Text} 
\begin{figure*}[h!]
    \centering
    \includegraphics[width=0.85\linewidth]{figures_paper/flexmore_super_figure_avg_1.png}
    \caption{
    Average downstream performance (Avg. Eq. \ref{eq:avg_score}) over 6 benchmarks (Eq. \ref{eq:eval_groups}).
    Relative performance ($\Delta\%$) across LoRA ranks for expert and FlexMoRE models.
    For each model family, $\Delta\%$ is computed relative to its corresponding baseline.
    Expert models are compared against their individual expert baselines, while FlexMoRE
    models are compared against the FlexOlmo baseline with the same number of active experts
    ($a2$, $a4$, $a7$).
    Extended results per benchmark are provided in Appendix A.
    }
    \label{fig:across_ranks}
\end{figure*}
\end{comment}

\subsection{Single-expert Results}\label{sec:result:comparing_experts}
Our main finding is that low-rank experts can be competitive and even outperform their full-size baseline.
Table \ref{sec:results:table:best_expert} shows the performance of the low-rank experts ($r \le 2^{11}$) against the FlexOlmo baseline experts (evaluated as mixture of the public base model and the respective expert). 
We observe consistent performance gains for low-rank experts, ranging from $0.73\%$ to $7.13\%$ for the Code, Creative Writing, Academic, and Reddit experts at ranks $2^9$, $2^4$, $2^6$, $2^{11}$, and $2^9$, respectively. In contrast, the Math expert exhibits a performance decrease of $1.34\%$, even when evaluated against the average at the highest meaningful rank ($2^{11}$).
Full results for all experts on all benchmarks across all tested ranks can be found in Appendix~\ref{sec:appendix-topk-experts}.
Crucially, no single LoRA rank is consistently best across all experts. Some evaluation groups exhibit flat performance across a wide range of ranks, whereas others favor higher-rank specialization.  To select the expert ranks for composing our FlexMoRE MoE models, we follow two strategies, one relying on experts' MC9 performance as a proxy and one relying on experts' average performance across all benchmarks. Details can be found in Appendix~\ref{sec:appendix-topk-experts}.

\begin{comment}
In brief, none of the experts have clear optimal ranks on any of the benchmarks, yet we do see some similar performance across experts on the GEN5 dataset. Interestingly, there, experts yield similar performance for lower ranks, whereas performance starts decreasing when $r>2^{10}$. This phenomenon is most clearly observed for the Reddit and Code experts. 
The Math expert benefits from higher ranks on all benchmarks except GEN5, where we observe a gradual decrease.
Overall, this analysis shows that optimal allocated capacity is strongly task-dependent, implying that rank heterogeneity might be beneficial.
\end{comment}

\begin{table*}[h]
\centering
\small
\sisetup{table-format=1.4}

\begin{tabular*}{\textwidth}{
@{\extracolsep{\fill}}
l
r|
*{6}{S}
|r
r
}
\toprule
\textbf{7x7B-1T Model} & \textbf{Params} &
{\textbf{MC9}} & {\textbf{GEN5}} & {\textbf{AGIEval}} & {\textbf{BBH}} &
{\textbf{MMLU}} & {\textbf{MMLU-Pro}} &
{\textbf{Avg.}} & {\textbf{$\Delta$(\%)}} \\
\midrule
FlexOlmo-a2 (baseline) & 33.27B 
& 0.6257 & 0.4508 & 0.3842 & \textbf{0.4469} & 0.5297 & 0.2415 & 0.4465 & -- \\[0.1ex]\cdashline{1-10}\addlinespace[0.5ex]
FlexMoRE-a2 (homogen. $r\!=\!2^9$) & 11.75B 
& 0.6722 & \textbf{0.4886} & 0.3976 & 0.3897 & 0.5483 & 0.2501 & 0.4577 & +2.53 \\
FlexMoRE-a2 (heterogen. all) & 14.84B 
& 0.6776 & 0.4780 & 0.4169 & 0.4195 & 0.5475 & 0.2537 & 0.4655 & +4.27 \\
FlexMoRE-a2 (heterogen. MC9) & \textbf{10.75B} 
& \textbf{0.6838} & 0.4868 & \textbf{0.4184} & 0.4269 & \textbf{0.5555} & \textbf{0.2543} & \textbf{0.4710} & \textbf{+5.49} \\
\midrule
FlexOlmo-a4 Baseline & 33.27B 
& 0.6709 & 0.4548 & 0.3999 & 0.4014 & 0.5518 & 0.2486 & 0.4546 & -- \\[0.1ex]\cdashline{1-10}\addlinespace[0.5ex]
FlexMoRE-a4 homogen. ($r\!=\!2^{10}$) & 16.21B 
& 0.6819 & 0.4784 & 0.3920 & 0.3887 & 0.5463 & 0.2487 & 0.4560 & +0.32 \\
FlexMoRE-a4 heterogen. (All) & 14.84B 
& 0.6896 & 0.4846 & 0.4121 & 0.4124 & \textbf{0.5557} & 0.2533 & 0.4679 & +2.94 \\
FlexMoRE-a4 heterogen. (MC9) & \textbf{10.75B} 
& \textbf{0.6936} & \textbf{0.4896} & \textbf{0.4140} & \textbf{0.4255} & 0.5516 & \textbf{0.2564} & \textbf{0.4718} & \textbf{+3.79} \\
\midrule
FlexOlmo-a7 Baseline & 33.27B 
& 0.6550 & 0.3614 & 0.3745 & 0.3560 & 0.5097 & 0.2212 & 0.4130 & -- \\[0.1ex]\cdashline{1-10}\addlinespace[0.5ex]
FlexMoRE-a7 homogen. ($r\!=\!2^{10}$) & 16.21B 
& 0.6733 & 0.4776 & 0.3680 & 0.3662 & 0.5209 & 0.2265 & 0.4388 & +6.25 \\
FlexMoRE-a7 heterogen. (All) & 14.84B 
& 0.6862 & 0.4834 & 0.4042 & 0.3861 & 0.5398 & 0.2420 & 0.4569 & +10.65 \\
FlexMoRE-a7 heterogen. (MC9) & \textbf{10.75B} 
& \textbf{0.6894} & \textbf{0.4896} & \textbf{0.4140} & \textbf{0.4255} & \textbf{0.5516} & \textbf{0.2564} & \textbf{0.4711} & \textbf{+14.08} \\

\bottomrule
\end{tabular*}
\caption{Results for the Mixture of Experts models  with $a2$, $a4$, and $a7$ active experts. 
The FlexOlmo baseline is a full-sized model without any low-rank adapters.
FlexMoRE homogeneous is the configuration with one full-size expert and all remaining six experts as low-rank adapters of the same rank (the ranks $r = 2^k,\; k \in \{0,\dots,11\}$, utilizing the rank that performs best on average across all benchmarks.
In FlexMoRE heterogenous (All), we selected the rank for each expert based on performance across all six benchmarks.
Finally, FlexMoRE heterogenous (MC9) is our best-performing proposed model, where we select the rank for each expert based on its performance on MC9. 
All scores are reported as mean performance. Relative improvements ($\Delta\%$) are computed against the corresponding baseline model.
}
\label{sec:results:table:best_flexmore}
\end{table*}

\subsection{Results of Mixture-of-Experts Models}\label{sec:downstream-baseline-analysis}

\begin{figure*}[t]
    \centering
    \includegraphics[width=0.8\linewidth]{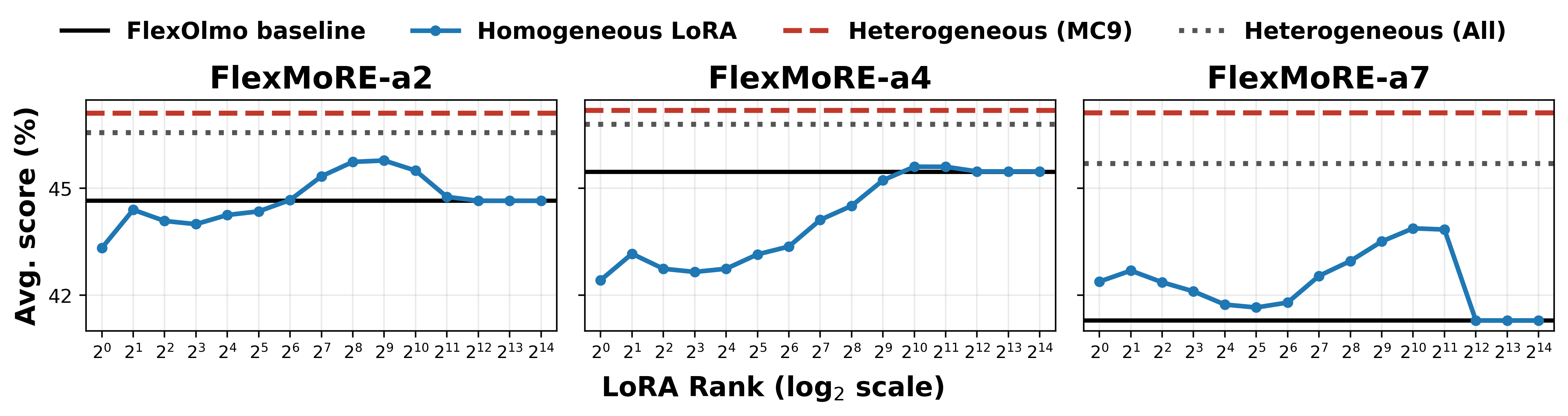}
    \caption{Unweighted average performance of FlexMoRE models with $2$, $4$, and $7$ active experts across six benchmarks.
    The solid curve shows performance under homogeneous post-hoc LoRA rank tuning. Dashed horizontal lines correspond to heterogeneous FlexMoRE compositions, with the dotted line indicating experts selected based on performance across all benchmarks (\emph{All}) and the dashed line experts selected using MC9 (\emph{MC9}).
    The FlexOlmo baseline is shown for reference.
    }
    \label{fig:results:a4}
\end{figure*}

Table \ref{sec:results:table:best_flexmore} reports the results of the MoE models (i.e., the public base model plus six experts) across the six eval groups. The experts' rank in our heterogeneous are either defined by individual expert's performance on MC9 or by average performance over all six benchmarks. We compare the heterogeneous model to a homogeneous FlexMoRE model with fixed rank across all experts and to the FlexOlmo baseline.

The heterogeneous experts outperforms both the FlexOlmo baseline and the homogeneous FlexMoRE models on average. Even more interestingly, the MC9- calibrated ranks perform better than the ranks calibrated across all six eval groups, performing consistently better on average performance \textit{Avg}. The MC9-calibrated rank-heterogeneous FlexMoRE models are globally best.

In Figure \ref{fig:results:a4} we report the 7-experts FlexMoRE with 2, 4, and 7 active experts per token. It is evident that while a homogeneous rank tuning can outperforms the FlexOlmo baseline with the right choice of rank, the heterogeneous FlexMoRE models, consistently outperform both the baseline and all rank-homogeneous FlexMoRE models.

Across all configurations, the FlexMoRE model comprising experts selected according to their performance on MC9 consistently outperforms selection based on the average performance across all six eval groups. This effect is most pronounced for configurations with very few (a2) or many (a7) active experts, while the difference is less pronounced but still non-negligible for a4 as shown in Table \ref{sec:results:table:best_flexmore} and in Figure \ref{fig:results:a4}.
Despite incorporating more task-specific information, the FlexMoRE model based on Avg favors capacity-heavy specialists whose performance gains occur at higher LoRA ranks, resulting in reduced effectiveness under fixed rank and routing budgets. In contrast, the FlexMoRE model based on MC9 performance implicitly prioritizes rank-efficient experts that deliver strong marginal gains at low to moderate ranks. This yield superior performance across eval groups. We provide further supporting material in Appendix \ref{sec:appendix-topk-experts}.

\paragraph{Task Diversity and Rank Efficiency} Domain-specialized experts perform best on the benchmarks aligned with their expected domains (e.g., Math on BBH, News on MMLU-Pro), typically peaking at higher ranks. This complementarity motivates heterogeneous post-hoc LoRA as a principled mechanism for uneven capacity allocation, achieving strong overall performance while substantially reducing parameter count and accelerator memory footprint compared to homogeneous rank tuning. Restricting rank calibration to unique experts yields qualitatively similar trends, confirming that results are not driven by multiple ranks of the same expert.

Generally, heterogeneous FlexMoRE models consistently outperform the homogeneous variants as well as the FlexOlmo baseline. All these gains come with considerable reductions in memory and computational requirements. For the single experts in our best-performing heterogeneous model based on MC9 performance, the memory requirements are reduced by 
$31.39\%$ for the Math expert, 
$95.71\%$ for the Creative Writing and Reddit experts, 
$97.85\%$ for the Code expert, 
$99.73\%$ for the Academic expert, and
$99.96\%$ for the News expert when compared to a full expert.

\subsection{Results of Rank Sensitivity Analysis}\label{sec:results:rank_sensitivity}
\begin{figure*}[t]
    \centering
    \includegraphics[width=\linewidth]{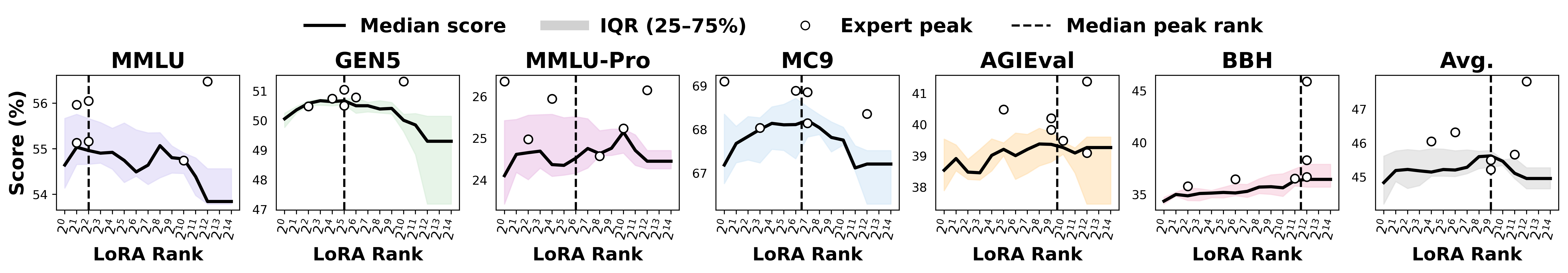}
    \caption{
    Typical log$_2$ LoRA rank at which experts achieve peak performance.
    For each expert $e$ and evaluation group $g$, the peak rank is computed directly from the observed scores after sorting by rank and resolving ties by selecting the lowest rank.
    Peak performance typically occurs at moderate ranks: for the aggregated average (Avg), the median peak is at $\log_2 r=9$ with IQR $[6.75,10.50]$ (i.e., ranks $\approx 2^7$–$2^{10}$).
    Knowledge-oriented benchmarks peak earlier (e.g., MMLU: median $\log_2 r=2$, IQR $[1.25,8.00]$; GEN5: median $\log_2 r=5$, IQR $[4.25,5.75]$),
    while reasoning-heavy benchmarks peak at substantially higher ranks (e.g., BBH: median $\log_2 r=11.5$, IQR $[7.25,12.00]$).
    }
    \label{fig:experts_rank_trends}
\end{figure*}

Figure~\ref{fig:experts_rank_trends} shows the rank–performance trends and benchmark-specific peak ranks. 
To quantify rank sensitivity, we fit a linear regression between log$_2$ rank and performance for each evaluation group (see Section~\ref{sub:rank-sensitivity-analysis}).  As shown in Figure~\ref{fig:experts_rank_trends},
reasoning-oriented benchmarks such as BBH exhibit consistently positive rank sensitivity in FlexMoRE 7x7B models, with slopes up to $0.0104$ and strong Pearson correlations ($r \approx 0.94$--$0.96$ for FlexMoRE $a2$ and $a4$), indicating a coherent and monotonic benefit from increased rank. In contrast, knowledge-oriented benchmarks such as GEN5 and MC9 frequently show weak or negative rank sensitivity (e.g., GEN5 slopes down to $-0.0042$ with $r \approx -0.86$), suggesting diminishing returns or overfitting at higher ranks. Detailed results on the rank sensitivity regression analysis can be found in Appendix~\ref{sec:appendix-rank-senstivity}.

Aggregating results across evaluation groups, FlexMoRE models exhibit positive typical rank sensitivity, with the strongest median effect observed at an intermediate number of active experts (FlexMoRE $a4$, median slope $0.0030$), followed by diminished gains for larger expert counts (FlexMoRE $a7$, median $0.0008$). In contrast, the most evaluated individually show near-zero or negative median rank sensitivity, that performance is unaffected by low-ranking. The exception is the math expert which benefits from higher ranks (median slope $0.0018$).
Details can be found in Appendix~\ref{sec:appendix-rank-senstivity}. 

\paragraph{Typical Peak Rank Per Task}
For each evaluation group, we summarize the distribution of $\log_2 r^*_{e,g}$ across experts using the median and interquartile range.
Results show that peak performance typically occurs at moderate ranks rather than at the maximum tested rank as shown in Figure~\ref{fig:experts_rank_trends}.
With respect to average performance, the median peak occurs at $\log_2 r = 9$ with an interquartile range of $[6.75, 10.5]$, corresponding approximately to ranks $2^7$--$2^{10}$.
Knowledge-oriented benchmarks peak substantially earlier (e.g., MMLU: median $\log_2 r = 2$), whereas reasoning-heavy benchmarks peak later (e.g., BBH: median $\log_2 r = 11.5$).

\paragraph{Expert-level Heterogeneity}
Aggregated rank trends mask substantial heterogeneity across individual experts.
To isolate expert-specific effects, we repeat the regression analysis independently for each expert using the aggregated average score across evaluation groups.
While some experts (most notably the Math expert) exhibit a strong and stable positive relationship between rank and performance, several experts show flat or negative slopes.
This heterogeneity explains the weak average rank sensitivity observed for expert-only models and motivates a per-expert analysis of rank effects (see Appendix~\ref{sec:appendix-rank-senstivity}).

\section{Discussion}
Our experiments demonstrated that the optimal rank of an expert is task-specific.
We show that MoEs with full size experts are suboptimal and limited in their scaling by the scarce accelerator memory.  We extended FlexOlmo to FlexMoRE, introducing low rank experts that enable researchers and practitioners to scale and collaborate on expert models without having to train full-size experts. These contributions come without compromise to overall model performance and contribute to the democratization of LLM development.   

\paragraph{Task-dependency of optimal ranks}
In Section \ref{sec:result:comparing_experts} we investigated the optimal rank for each expert across tasks. We demonstrated that five out of six experts can be post-hoc low-ranked yielding superior downstream task performance. The math expert proves sensitive to lower rank and exhibits a slight decrease in performance. Our analysis suggests that knowledge-heavy datasets thrive in lower ranks while reasoning-based tasks require higher ranks. This is not entirely surprising as reasoning tasks generally are more complex and known to require more capacity.

\paragraph{Rank heterogeneity and expert contribution}
Our expert-level analysis shows that peak performance occurs at widely varying LoRA ranks across domains. Crucially, high peak performance does not imply high contribution within a mixture. Instead, contribution is governed by the interaction between rank efficiency, routing frequency, and task coverage.
This explains why experts that dominate individual benchmarks at high ranks do not necessarily improve combined models, while experts with modest peak scores but early gains contribute disproportionately to overall performance.

\paragraph{Global rank trends and implications}
Jointly, these results show that LoRA ranks matter but that their impact is task-dependent and complex.
While increasing rank can improve performance, gains saturate well before the maximum tested rank for most experts and tasks.
There is no single optimal rank across benchmarks; instead, effective rank allocation depends strongly on task characteristics.
In practice, ranks beyond approximately $\log_2 r \approx 9$--$10$ yield diminishing returns for most experts, suggesting that LoRA ranks should be allocated selectively rather than scaled uniformly. This point is heavily underscored by our results from rank-heterogenous models, which consistently outperform homogeneous ones regardless of the rank of the homogeneous model.

\paragraph{Expert Rank Selection}
We have studied two possible approaches to determine the optimal rank of each expert in a FlexMoRE model. In the first option, we use the MC9 benchmark as a proxy. The second option takes the global average performance of an expert across all six benchmarks. Note that in both cases, the selection happens based on the performance of individual experts, not on the performance of the entire mixture.
Interestingly, the MC9 rank selection strategy turns out favorably, indicating that multiple-choice tasks may be a sufficient proxy for expert performance, while other benchmarks may dilute the selection process as they, for instance, lead to higher than necessary ranks. 
In the final mixture, high-rank experts may counterbalance low-rank experts.

\paragraph{Limitations} Our work comes with several limitations. First, we did not apply router-tuning after post-hoc LoRA extraction. The effect of router-tuning was marginal in FlexOlmo, and the clear expectation is that the performance would only be further improved with router tuning. Second, while our architecture allows having multiple full-size experts, transitive chains of low rank experts, and LoRA training from scratch, in this paper, we limited ourselves to post-hoc LoRA extraction from pre-trained FlexOLMo models. This has allowed us to conduct a comprehensive set of experiments to determine rank-sensitivity under fixed conditions.

\section{Conclusion}
We introducedd FlexMoRE, a flexible mixture-of-experts architecture supporting rank-heterogeneous low-rank adapters for federated LLM training alongside full-size experts. Empirically, we demonstrated that low rank experts can improve performance and reduce parameters by 67\%. Crucially, we find that optimal expert rank is task-dependent: reasoning-heavy benchmarks require higher ranks than knowledge-oriented tasks. These findings suggest that rank should be allocated selectively rather than uniformly to achieve substantial memory savings without sacrificing performance.

\paragraph{Future Work}
First, future work could further investigate rank selection strategies.  Second, future work could investigate the optimal combination of full-size experts, e.g., one per language in a multilingual setup alongside one low-rank expert per domain.
Last, future work could investigate introducing even more rank heterogeneity by distributing capacity selectively among layers (cf.\ L1RA~\cite{singh2025l1ra}).

\FloatBarrier

\bibliographystyle{named}
\bibliography{ijcai26}

@article{anil2023palm,
  title={Palm 2 technical report},
  author={Anil, Rohan and Dai, Andrew M and Firat, Orhan and Johnson, Melvin and Lepikhin, Dmitry and Passos, Alexandre and Shakeri, Siamak and Taropa, Emanuel and Bailey, Paige and Chen, Zhifeng and others},
  journal={arXiv preprint arXiv:2305.10403},
  year={2023}
}

@article{chowdhery2023palm,
  title={Palm: Scaling language modeling with pathways},
  author={Chowdhery, Aakanksha and Narang, Sharan and Devlin, Jacob and Bosma, Maarten and Mishra, Gaurav and Roberts, Adam and Barham, Paul and Chung, Hyung Won and Sutton, Charles and Gehrmann, Sebastian and others},
  journal={Journal of Machine Learning Research},
  volume={24},
  number={240},
  pages={1--113},
  year={2023}
}

@misc{clark2018thinksolvedquestionanswering,
      title={Think you have Solved Question Answering? Try ARC, the AI2 Reasoning Challenge}, 
      author={Peter Clark and Isaac Cowhey and Oren Etzioni and Tushar Khot and Ashish Sabharwal and Carissa Schoenick and Oyvind Tafjord},
      year={2018},
      eprint={1803.05457},
      archivePrefix={arXiv},
      primaryClass={cs.AI},
      url={https://arxiv.org/abs/1803.05457}, 
}

@misc{clark2019boolqexploringsurprisingdifficulty,
      title={BoolQ: Exploring the Surprising Difficulty of Natural Yes/No Questions}, 
      author={Christopher Clark and Kenton Lee and Ming-Wei Chang and Tom Kwiatkowski and Michael Collins and Kristina Toutanova},
      year={2019},
      eprint={1905.10044},
      archivePrefix={arXiv},
      primaryClass={cs.CL},
      url={https://arxiv.org/abs/1905.10044}, 
}

@misc{reddy2019coqaconversationalquestionanswering,
      title={CoQA: A Conversational Question Answering Challenge}, 
      author={Siva Reddy and Danqi Chen and Christopher D. Manning},
      year={2019},
      eprint={1808.07042},
      archivePrefix={arXiv},
      primaryClass={cs.CL},
      url={https://arxiv.org/abs/1808.07042}, 
}

@misc{zellers2019hellaswagmachinereallyfinish,
      title={HellaSwag: Can a Machine Really Finish Your Sentence?}, 
      author={Rowan Zellers and Ari Holtzman and Yonatan Bisk and Ali Farhadi and Yejin Choi},
      year={2019},
      eprint={1905.07830},
      archivePrefix={arXiv},
      primaryClass={cs.CL},
      url={https://arxiv.org/abs/1905.07830}, 
}

@misc{mihaylov2018suitarmorconductelectricity,
      title={Can a Suit of Armor Conduct Electricity? A New Dataset for Open Book Question Answering}, 
      author={Todor Mihaylov and Peter Clark and Tushar Khot and Ashish Sabharwal},
      year={2018},
      eprint={1809.02789},
      archivePrefix={arXiv},
      primaryClass={cs.CL},
      url={https://arxiv.org/abs/1809.02789}, 
}

@misc{bisk2019piqareasoningphysicalcommonsense,
      title={PIQA: Reasoning about Physical Commonsense in Natural Language}, 
      author={Yonatan Bisk and Rowan Zellers and Ronan Le Bras and Jianfeng Gao and Yejin Choi},
      year={2019},
      eprint={1911.11641},
      archivePrefix={arXiv},
      primaryClass={cs.CL},
      url={https://arxiv.org/abs/1911.11641}, 
}

@misc{sap2019socialiqacommonsensereasoningsocial,
      title={SocialIQA: Commonsense Reasoning about Social Interactions}, 
      author={Maarten Sap and Hannah Rashkin and Derek Chen and Ronan LeBras and Yejin Choi},
      year={2019},
      eprint={1904.09728},
      archivePrefix={arXiv},
      primaryClass={cs.CL},
      url={https://arxiv.org/abs/1904.09728}, 
}

@misc{sakaguchi2019winograndeadversarialwinogradschema,
      title={WinoGrande: An Adversarial Winograd Schema Challenge at Scale}, 
      author={Keisuke Sakaguchi and Ronan Le Bras and Chandra Bhagavatula and Yejin Choi},
      year={2019},
      eprint={1907.10641},
      archivePrefix={arXiv},
      primaryClass={cs.CL},
      url={https://arxiv.org/abs/1907.10641}, 
}

@misc{rajpurkar2016squad100000questionsmachine,
      title={SQuAD: 100,000+ Questions for Machine Comprehension of Text}, 
      author={Pranav Rajpurkar and Jian Zhang and Konstantin Lopyrev and Percy Liang},
      year={2016},
      eprint={1606.05250},
      archivePrefix={arXiv},
      primaryClass={cs.CL},
      url={https://arxiv.org/abs/1606.05250}, 
}

@misc{joshi2017triviaqalargescaledistantly,
      title={TriviaQA: A Large Scale Distantly Supervised Challenge Dataset for Reading Comprehension}, 
      author={Mandar Joshi and Eunsol Choi and Daniel S. Weld and Luke Zettlemoyer},
      year={2017},
      eprint={1705.03551},
      archivePrefix={arXiv},
      primaryClass={cs.CL},
      url={https://arxiv.org/abs/1705.03551}, 
}

@misc{dua2019dropreadingcomprehensionbenchmark,
      title={DROP: A Reading Comprehension Benchmark Requiring Discrete Reasoning Over Paragraphs}, 
      author={Dheeru Dua and Yizhong Wang and Pradeep Dasigi and Gabriel Stanovsky and Sameer Singh and Matt Gardner},
      year={2019},
      eprint={1903.00161},
      archivePrefix={arXiv},
      primaryClass={cs.CL},
      url={https://arxiv.org/abs/1903.00161}, 
}

@misc{zhong2023agievalhumancentricbenchmarkevaluating,
      title={AGIEval: A Human-Centric Benchmark for Evaluating Foundation Models}, 
      author={Wanjun Zhong and Ruixiang Cui and Yiduo Guo and Yaobo Liang and Shuai Lu and Yanlin Wang and Amin Saied and Weizhu Chen and Nan Duan},
      year={2023},
      eprint={2304.06364},
      archivePrefix={arXiv},
      primaryClass={cs.CL},
      url={https://arxiv.org/abs/2304.06364}, 
}

@misc{suzgun2022challengingbigbenchtaskschainofthought,
      title={Challenging BIG-Bench Tasks and Whether Chain-of-Thought Can Solve Them}, 
      author={Mirac Suzgun and Nathan Scales and Nathanael Schärli and Sebastian Gehrmann and Yi Tay and Hyung Won Chung and Aakanksha Chowdhery and Quoc V. Le and Ed H. Chi and Denny Zhou and Jason Wei},
      year={2022},
      eprint={2210.09261},
      archivePrefix={arXiv},
      primaryClass={cs.CL},
      url={https://arxiv.org/abs/2210.09261}, 
}

@misc{hendrycks2021measuringmassivemultitasklanguage,
      title={Measuring Massive Multitask Language Understanding}, 
      author={Dan Hendrycks and Collin Burns and Steven Basart and Andy Zou and Mantas Mazeika and Dawn Song and Jacob Steinhardt},
      year={2021},
      eprint={2009.03300},
      archivePrefix={arXiv},
      primaryClass={cs.CY},
      url={https://arxiv.org/abs/2009.03300}, 
}

@article{Wang2024MMLUProAM,
  title={MMLU-Pro: A More Robust and Challenging Multi-Task Language Understanding Benchmark},
  author={Yubo Wang and Xueguang Ma and Ge Zhang and Yuansheng Ni and Abhranil Chandra and Shiguang Guo and Weiming Ren and Aaran Arulraj and Xuan He and Ziyan Jiang and Tianle Li and Max W.F. Ku and Kai Wang and Alex Zhuang and Rongqi "Richard" Fan and Xiang Yue and Wenhu Chen},
  journal={ArXiv},
  year={2024},
  volume={abs/2406.01574},
  url={https://api.semanticscholar.org/CorpusID:270210486}
}

@article{kwiatkowski-etal-2019-natural,
    title = "Natural Questions: A Benchmark for Question Answering Research",
    author = "Kwiatkowski, Tom  and
      Palomaki, Jennimaria  and
      Redfield, Olivia  and
      Collins, Michael  and
      Parikh, Ankur  and
      Alberti, Chris  and
      Epstein, Danielle  and
      Polosukhin, Illia  and
      Devlin, Jacob  and
      Lee, Kenton  and
      Toutanova, Kristina  and
      Jones, Llion  and
      Kelcey, Matthew  and
      Chang, Ming-Wei  and
      Dai, Andrew M.  and
      Uszkoreit, Jakob  and
      Le, Quoc  and
      Petrov, Slav",
    editor = "Lee, Lillian  and
      Johnson, Mark  and
      Roark, Brian  and
      Nenkova, Ani",
    journal = "Transactions of the Association for Computational Linguistics",
    volume = "7",
    year = "2019",
    address = "Cambridge, MA",
    publisher = "MIT Press",
    url = "https://aclanthology.org/Q19-1026/",
    doi = "10.1162/tacl_a_00276",
    pages = "452--466"
}

@misc{xie2025dflmoedecentralizedfederatedlearning,
      title={dFLMoE: Decentralized Federated Learning via Mixture of Experts for Medical Data Analysis}, 
      author={Luyuan Xie and Tianyu Luan and Wenyuan Cai and Guochen Yan and Zhaoyu Chen and Nan Xi and Yuejian Fang and Qingni Shen and Zhonghai Wu and Junsong Yuan},
      year={2025},
      eprint={2503.10412},
      archivePrefix={arXiv},
      primaryClass={cs.LG},
      url={https://arxiv.org/abs/2503.10412}, 
}

@misc{vasani2025phloradatafreeposthoclowrank,
      title={PHLoRA: data-free Post-hoc Low-Rank Adapter extraction from full-rank checkpoint}, 
      author={Bhoomit Vasani and Jack FitzGerald and Anjie Fang and Sushmit Vaish},
      year={2025},
      eprint={2509.10971},
      archivePrefix={arXiv},
      primaryClass={cs.LG},
      url={https://arxiv.org/abs/2509.10971}, 
}

@misc{mu2025comprehensivesurveymixtureofexpertsalgorithms,
      title={A Comprehensive Survey of Mixture-of-Experts: Algorithms, Theory, and Applications}, 
      author={Siyuan Mu and Sen Lin},
      year={2025},
      eprint={2503.07137},
      archivePrefix={arXiv},
      primaryClass={cs.LG},
      url={https://arxiv.org/abs/2503.07137}, 
}

@misc{zhou2022mixtureofexpertsexpertchoicerouting,
      title={Mixture-of-Experts with Expert Choice Routing}, 
      author={Yanqi Zhou and Tao Lei and Hanxiao Liu and Nan Du and Yanping Huang and Vincent Zhao and Andrew Dai and Zhifeng Chen and Quoc Le and James Laudon},
      year={2022},
      eprint={2202.09368},
      archivePrefix={arXiv},
      primaryClass={cs.LG},
      url={https://arxiv.org/abs/2202.09368}, 
}

@misc{rajbhandari2022deepspeedmoeadvancingmixtureofexpertsinference,
      title={DeepSpeed-MoE: Advancing Mixture-of-Experts Inference and Training to Power Next-Generation AI Scale}, 
      author={Samyam Rajbhandari and Conglong Li and Zhewei Yao and Minjia Zhang and Reza Yazdani Aminabadi and Ammar Ahmad Awan and Jeff Rasley and Yuxiong He},
      year={2022},
      eprint={2201.05596},
      archivePrefix={arXiv},
      primaryClass={cs.LG},
      url={https://arxiv.org/abs/2201.05596}, 
}

@inproceedings{Wang_2025, series={ACM MOBICOM ’25},
   title={D2MoE: Dual Routing and Dynamic Scheduling for Efficient On-Device MoE-based LLM Serving},
   url={http://dx.doi.org/10.1145/3680207.3723493},
   DOI={10.1145/3680207.3723493},
   booktitle={Proceedings of the 31st Annual International Conference on Mobile Computing and Networking},
   publisher={ACM},
   author={Wang, Haodong and Zhou, Qihua and Hong, Zicong and Guo, Song},
   year={2025},
   month=nov, pages={574–588},
   collection={ACM MOBICOM ’25} }

@misc{zadouri2023pushingmixtureexpertslimit,
      title={Pushing Mixture of Experts to the Limit: Extremely Parameter Efficient MoE for Instruction Tuning}, 
      author={Ted Zadouri and Ahmet Üstün and Arash Ahmadian and Beyza Ermiş and Acyr Locatelli and Sara Hooker},
      year={2023},
      eprint={2309.05444},
      archivePrefix={arXiv},
      primaryClass={cs.CL},
      url={https://arxiv.org/abs/2309.05444}, 
}

@misc{cao2024moelightninghighthroughputmoeinference,
      title={MoE-Lightning: High-Throughput MoE Inference on Memory-constrained GPUs}, 
      author={Shiyi Cao and Shu Liu and Tyler Griggs and Peter Schafhalter and Xiaoxuan Liu and Ying Sheng and Joseph E. Gonzalez and Matei Zaharia and Ion Stoica},
      year={2024},
      eprint={2411.11217},
      archivePrefix={arXiv},
      primaryClass={cs.DC},
      url={https://arxiv.org/abs/2411.11217}, 
}

@misc{zhao2025puzzlemoeefficientcompressionlarge,
      title={PuzzleMoE: Efficient Compression of Large Mixture-of-Experts Models via Sparse Expert Merging and Bit-packed inference}, 
      author={Yushu Zhao and Zheng Wang and Minjia Zhang},
      year={2025},
      eprint={2511.04805},
      archivePrefix={arXiv},
      primaryClass={cs.LG},
      url={https://arxiv.org/abs/2511.04805}, 
}

@misc{hu2023llmadaptersadapterfamilyparameterefficient,
      title={LLM-Adapters: An Adapter Family for Parameter-Efficient Fine-Tuning of Large Language Models}, 
      author={Zhiqiang Hu and Lei Wang and Yihuai Lan and Wanyu Xu and Ee-Peng Lim and Lidong Bing and Xing Xu and Soujanya Poria and Roy Ka-Wei Lee},
      year={2023},
      eprint={2304.01933},
      archivePrefix={arXiv},
      primaryClass={cs.CL},
      url={https://arxiv.org/abs/2304.01933}, 
}

@misc{zhuang2025ldmolelearnabledynamicrouting,
      title={LD-MoLE: Learnable Dynamic Routing for Mixture of LoRA Experts}, 
      author={Yuan Zhuang and Yi Shen and Yuexin Bian and Qing Su and Shihao Ji and Yuanyuan Shi and Fei Miao},
      year={2025},
      eprint={2509.25684},
      archivePrefix={arXiv},
      primaryClass={cs.CL},
      url={https://arxiv.org/abs/2509.25684}, 
}

@misc{kunwar2025ttloramoeunifyingparameterefficient,
      title={TT-LoRA MoE: Unifying Parameter-Efficient Fine-Tuning and Sparse Mixture-of-Experts}, 
      author={Pradip Kunwar and Minh N. Vu and Maanak Gupta and Mahmoud Abdelsalam and Manish Bhattarai},
      year={2025},
      eprint={2504.21190},
      archivePrefix={arXiv},
      primaryClass={cs.LG},
      url={https://arxiv.org/abs/2504.21190}, 
}

@misc{cao2025moaheterogeneousmixtureadapters,
      title={MoA: Heterogeneous Mixture of Adapters for Parameter-Efficient Fine-Tuning of Large Language Models}, 
      author={Jie Cao and Tianwei Lin and Hongyang He and Rolan Yan and Wenqiao Zhang and Juncheng Li and Dongping Zhang and Siliang Tang and Yueting Zhuang},
      year={2025},
      eprint={2506.05928},
      archivePrefix={arXiv},
      primaryClass={cs.CL},
      url={https://arxiv.org/abs/2506.05928}, 
}

@inproceedings{wang2022adamix,
  title={Adamix: Mixture-of-adaptations for parameter-efficient model tuning},
  author={Wang, Yaqing and Agarwal, Sahaj and Mukherjee, Subhabrata and Liu, Xiaodong and Gao, Jing and Hassan, Ahmed and Gao, Jianfeng},
  booktitle={Proceedings of the 2022 Conference on Empirical Methods in Natural Language Processing},
  pages={5744--5760},
  year={2022}
}

@misc{ji2025lmoeendtoendtraininglightweight,
      title={L-MoE: End-to-End Training of a Lightweight Mixture of Low-Rank Adaptation Experts}, 
      author={Shihao Ji and Zihui Song},
      year={2025},
      eprint={2510.17898},
      archivePrefix={arXiv},
      primaryClass={cs.LG},
      url={https://arxiv.org/abs/2510.17898}, 
}

@misc{li2024mixloraenhancinglargelanguage,
      title={MixLoRA: Enhancing Large Language Models Fine-Tuning with LoRA-based Mixture of Experts}, 
      author={Dengchun Li and Yingzi Ma and Naizheng Wang and Zhengmao Ye and Zhiyuan Cheng and Yinghao Tang and Yan Zhang and Lei Duan and Jie Zuo and Cal Yang and Mingjie Tang},
      year={2024},
      eprint={2404.15159},
      archivePrefix={arXiv},
      primaryClass={cs.CL},
      url={https://arxiv.org/abs/2404.15159}, 
}

@misc{zou2025flyloraboostingtaskdecoupling,
      title={FlyLoRA: Boosting Task Decoupling and Parameter Efficiency via Implicit Rank-Wise Mixture-of-Experts}, 
      author={Heming Zou and Yunliang Zang and Wutong Xu and Yao Zhu and Xiangyang Ji},
      year={2025},
      eprint={2510.08396},
      archivePrefix={arXiv},
      primaryClass={cs.LG},
      url={https://arxiv.org/abs/2510.08396}, 
}

@misc{shi2025flexolmoopenlanguagemodels,
      title={FlexOlmo: Open Language Models for Flexible Data Use}, 
      author={Weijia Shi and Akshita Bhagia and Kevin Farhat and Niklas Muennighoff and Pete Walsh and Jacob Morrison and Dustin Schwenk and Shayne Longpre and Jake Poznanski and Allyson Ettinger and Daogao Liu and Margaret Li and Dirk Groeneveld and Mike Lewis and Wen-tau Yih and Luca Soldaini and Kyle Lo and Noah A. Smith and Luke Zettlemoyer and Pang Wei Koh and Hannaneh Hajishirzi and Ali Farhadi and Sewon Min},
      year={2025},
      eprint={2507.07024},
      archivePrefix={arXiv},
      primaryClass={cs.CL},
      url={https://arxiv.org/abs/2507.07024}, 
}

@misc{yi2024pfedmoedatalevelpersonalizationmixture,
      title={pFedMoE: Data-Level Personalization with Mixture of Experts for Model-Heterogeneous Personalized Federated Learning}, 
      author={Liping Yi and Han Yu and Chao Ren and Heng Zhang and Gang Wang and Xiaoguang Liu and Xiaoxiao Li},
      year={2024},
      eprint={2402.01350},
      archivePrefix={arXiv},
      primaryClass={cs.LG},
      url={https://arxiv.org/abs/2402.01350}, 
}

@Article{bdcc9060147,
AUTHOR = {Pahune, Saurabh and Akhtar, Zahid and Mandapati, Venkatesh and Siddique, Kamran},
TITLE = {The Importance of AI Data Governance in Large Language Models},
JOURNAL = {Big Data and Cognitive Computing},
VOLUME = {9},
YEAR = {2025},
NUMBER = {6},
ARTICLE-NUMBER = {147},
URL = {https://www.mdpi.com/2504-2289/9/6/147},
ISSN = {2504-2289},
DOI = {10.3390/bdcc9060147}
}

@article{Abishek_2025,
   title={Data and AI governance: Promoting equity, ethics, and fairness in large language models},
   volume={6},
   url={http://dx.doi.org/10.38105/spr.1sn574k4lp},
   DOI={10.38105/spr.1sn574k4lp},
   journal={MIT Science Policy Review},
   publisher={MIT Science Policy Review},
   author={Abishek, Alok and Erickson, Lisa and Bandopadhyay, Tushar},
   year={2025},
   month=aug, pages={139–146} }

@article{fedus2022switch,
  title={Switch transformers: Scaling to trillion parameter models with simple and efficient sparsity},
  author={Fedus, William and Zoph, Barret and Shazeer, Noam},
  journal={Journal of Machine Learning Research},
  volume={23},
  number={120},
  pages={1--39},
  year={2022}
}

@article{liu2024deepseek,
  title={Deepseek-v3 technical report},
  author={Liu, Aixin and Feng, Bei and Xue, Bing and Wang, Bingxuan and Wu, Bochao and Lu, Chengda and Zhao, Chenggang and Deng, Chengqi and Zhang, Chenyu and Ruan, Chong and others},
  journal={arXiv preprint arXiv:2412.19437},
  year={2024}
}

@article{hu2022lora,
  title={Lora: Low-rank adaptation of large language models.},
  author={Hu, Edward J and Shen, Yelong and Wallis, Phillip and Allen-Zhu, Zeyuan and Li, Yuanzhi and Wang, Shean and Wang, Lu and Chen, Weizhu and others},
  journal={ICLR},
  volume={1},
  number={2},
  pages={3},
  year={2022}
}

@inproceedings{singh2025l1ra,
  title={L1RA: Dynamic Rank Assignment in LoRA Fine-Tuning},
  author={Singh, Raul and Brunello, Nicol{\`o} and Scotti, Vincenzo and Carman, Mark},
  booktitle={Proceedings of the 8th International Conference on Natural Language and Speech Processing (ICNLSP-2025)},
  pages={360--373},
  year={2025}
}

\appendix

\clearpage

\section{Top-$k$ Experts Benchmark Performance Analysis}
\label{sec:appendix-topk-experts}

\begin{figure*}[htb]
    \centering
    \includegraphics[width=1.0\textwidth]{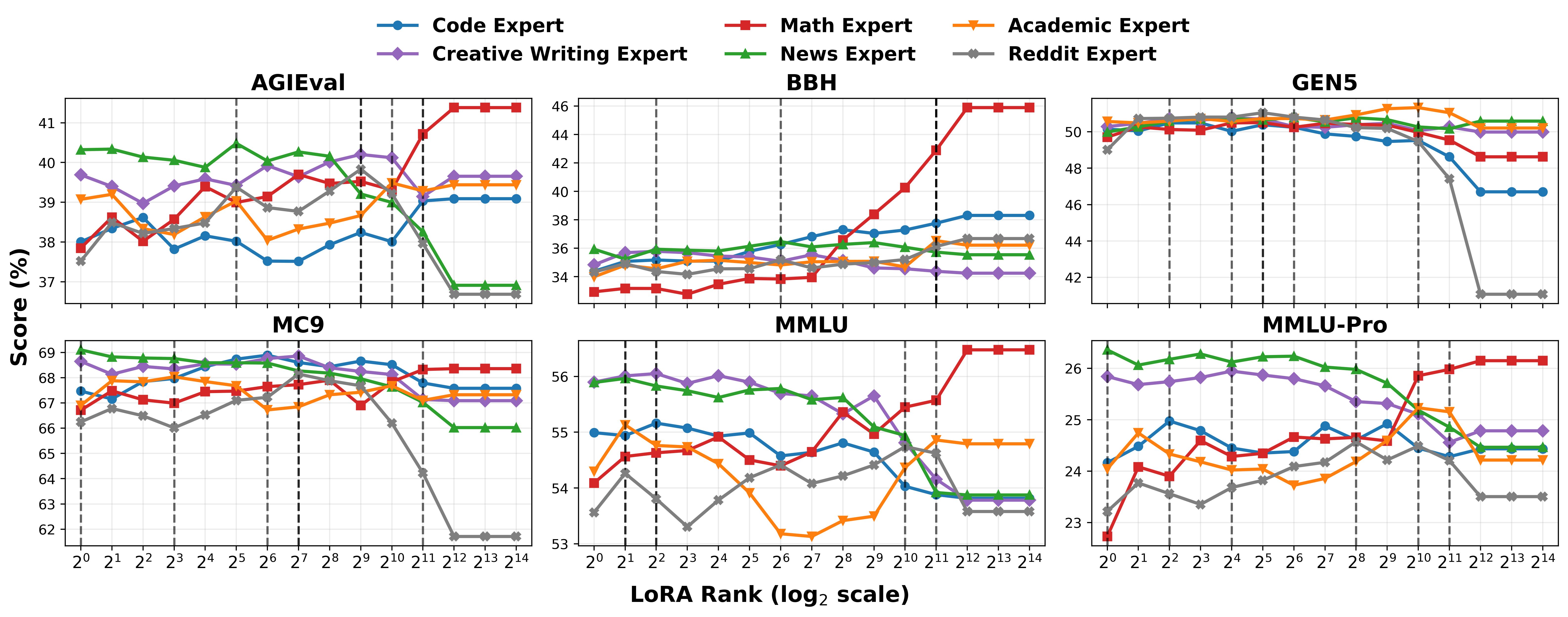}
    \caption{
    Performance across all six benchmarks illustrating expert specialization and rank sensitivity.
    AGIEval emphasizes knowledge-intensive and academic-style tasks, where the experts like perform strongly.
    BBH focuses on structured reasoning, favoring the \emph{Code} and \emph{Math} experts.
    GEN5 captures open-ended and generative abilities, where the \emph{Creative Writing} and \emph{Reddit} experts are most competitive.
    MC9 evaluates mixed-task performance and favors rank-efficient generalist experts with broad cross-domain utility.
    }
    \label{fig:results:experts}
\end{figure*}

Figure~\ref{fig:results:experts} provides the results for all experts on all benchmarks across all tested ranks.

To contextualize expert selection behavior across benchmarks, we report two complementary analyses.

First, we list the top-$k$ ranked expert models per benchmark, where each entry corresponds to a specific expert–rank pair.

Second, we report a top-$k$ analysis restricted to unique experts, where for each expert the best-performing LoRA rank (rank $\leq 2^{11}$) is selected.
In both cases, rankings are computed independently per benchmark, with ties resolved by selecting the lowest LoRA rank.

Table~\ref{tab:topk_models} reports the top-$k$ expert models per benchmark without enforcing uniqueness across experts.
This view highlights which expert–rank combinations achieve the highest scores on each benchmark, allowing multiple ranks of the same expert to appear.
The analysis reflects peak standalone performance under the rank constraint, independent of mixture composition.

Table~\ref{tab:topk_unique_experts} reports a complementary analysis restricted to unique experts.
For each benchmark, each expert appears at most once, using its best-performing LoRA rank within the rank constraint.
This view isolates expert-level performance independent of rank multiplicity and facilitates comparison across domains.

\begin{table*}[h]
\centering
\small
\sisetup{table-format=1.4}

\begin{tabular*}{\textwidth}{
@{\extracolsep{\fill}}
l
S
l
S
S
}
\toprule
\textbf{Benchmark} & \textbf{Ranking} & \textbf{Model} & \textbf{LoRA rank} & \textbf{Score} \\
\midrule
\multicolumn{5}{l}{\textbf{\textit{MC9}}} \\
MC9 & 1 & Flex-news-2x7B-1T-r1 & 1 & 0.6910 \\
MC9 & 2 & Flex-code-2x7B-1T-r64 & 64 & 0.6889 \\
MC9 & 3 & Flex-creative-2x7B-1T-r128 & 128 & 0.6886 \\
MC9 & 4 & Flex-news-2x7B-1T-r2 & 2 & 0.6882 \\
MC9 & 5 & Flex-news-2x7B-1T-r4 & 4 & 0.6878 \\
MC9 & 6 & Flex-news-2x7B-1T-r8 & 8 & 0.6876 \\
\midrule
\multicolumn{5}{l}{\textbf{\textit{GEN5}}}\\
GEN5 & 1 & Flex-pes2o-2x7B-1T-r1024 & 1024 & 0.5132 \\
GEN5 & 2 & Flex-pes2o-2x7B-1T-r512 & 512 & 0.5126 \\
GEN5 & 3 & Flex-reddit-2x7B-1T-r32 & 32 & 0.5104 \\
GEN5 & 4 & Flex-pes2o-2x7B-1T-r2048 & 2048 & 0.5104 \\
GEN5 & 5 & Flex-pes2o-2x7B-1T-r256 & 256 & 0.5092 \\
GEN5 & 6 & Flex-reddit-2x7B-1T-r8 & 8 & 0.5080 \\
\midrule
\multicolumn{5}{l}{\textbf{\textit{AGIEval}}}\\
AGIEval & 1 & Flex-math-2x7B-1T-r2048 & 2048 & 0.4072 \\
AGIEval & 2 & Flex-news-2x7B-1T-r32 & 32 & 0.4048 \\
AGIEval & 3 & Flex-news-2x7B-1T-r2 & 2 & 0.4034 \\
AGIEval & 4 & Flex-news-2x7B-1T-r1 & 1 & 0.4032 \\
AGIEval & 5 & Flex-news-2x7B-1T-r128 & 128 & 0.4027 \\
AGIEval & 6 & Flex-creative-2x7B-1T-r512 & 512 & 0.4020 \\
\midrule
\multicolumn{5}{l}{\textbf{\textit{BBH}}}\\
BBH & 1 & Flex-math-2x7B-1T-r2048 & 2048 & 0.4289 \\
BBH & 2 & Flex-math-2x7B-1T-r1024 & 1024 & 0.4026 \\
BBH & 3 & Flex-math-2x7B-1T-r512 & 512 & 0.3839 \\
BBH & 4 & Flex-code-2x7B-1T-r2048 & 2048 & 0.3776 \\
BBH & 5 & Flex-code-2x7B-1T-r256 & 256 & 0.3730 \\
BBH & 6 & Flex-code-2x7B-1T-r1024 & 1024 & 0.3728 \\
\midrule
\multicolumn{5}{l}{\textbf{\textit{MMLU}}}\\
MMLU & 1 & Flex-creative-2x7B-1T-r4 & 4 & 0.5605 \\
MMLU & 2 & Flex-creative-2x7B-1T-r16 & 16 & 0.5601 \\
MMLU & 3 & Flex-creative-2x7B-1T-r2 & 2 & 0.5601 \\
MMLU & 4 & Flex-news-2x7B-1T-r2 & 2 & 0.5596 \\
MMLU & 5 & Flex-creative-2x7B-1T-r1 & 1 & 0.5590 \\
MMLU & 6 & Flex-creative-2x7B-1T-r32 & 32 & 0.5589 \\
\midrule
\multicolumn{5}{l}{\textbf{\textit{MMLU-Pro}}}\\
MMLU-Pro & 1 & Flex-news-2x7B-1T-r1 & 1 & 0.2636 \\
MMLU-Pro & 2 & Flex-news-2x7B-1T-r8 & 8 & 0.2628 \\
MMLU-Pro & 3 & Flex-news-2x7B-1T-r64 & 64 & 0.2624 \\
MMLU-Pro & 4 & Flex-news-2x7B-1T-r32 & 32 & 0.2623 \\
MMLU-Pro & 5 & Flex-news-2x7B-1T-r4 & 4 & 0.2618 \\
MMLU-Pro & 6 & Flex-news-2x7B-1T-r16 & 16 & 0.2612 \\
\bottomrule
\end{tabular*}
\caption{
Top-6 ranked expert models per benchmark, restricted to LoRA ranks $\leq 2^{11}$. Ties are resolved by selecting the lowest rank.
Expert mapping: news (News), code (Code), creative (Creative Writing), pes2o (Academic), reddit (Reddit) and math (Math)
}
\label{tab:topk_models}
\end{table*}

\begin{table*}[h]
\centering
\small
\sisetup{table-format=1.4}

\begin{tabular*}{\textwidth}{
@{\extracolsep{\fill}}
l
S
l
S
S
}
\toprule
\textbf{Benchmark} & \textbf{Ranking} & \textbf{Model} & \textbf{LoRA rank} & \textbf{Score} \\
\midrule
\multicolumn{5}{l}{\textbf{\textit{MC9}}}\\
MC9 & 1 & Flex-news-2x7B-1T & 1 & 0.6910 \\
MC9 & 2 & Flex-code-2x7B-1T & 64 & 0.6889 \\
MC9 & 3 & Flex-creative-2x7B-1T & 128 & 0.6886 \\
MC9 & 4 & Flex-math-2x7B-1T & 2048 & 0.6832 \\
MC9 & 5 & Flex-reddit-2x7B-1T & 128 & 0.6814 \\
MC9 & 6 & Flex-pes2o-2x7B-1T & 8 & 0.6803 \\
\midrule
\multicolumn{5}{l}{\textbf{\textit{GEN5}}}\\
GEN5 & 1 & Flex-pes2o-2x7B-1T & 1024 & 0.5132 \\
GEN5 & 2 & Flex-reddit-2x7B-1T & 32 & 0.5104 \\
GEN5 & 3 & Flex-news-2x7B-1T & 64 & 0.5078 \\
GEN5 & 4 & Flex-creative-2x7B-1T & 16 & 0.5074 \\
GEN5 & 5 & Flex-math-2x7B-1T & 32 & 0.5050 \\
GEN5 & 6 & Flex-code-2x7B-1T & 4 & 0.5048 \\
\midrule
\multicolumn{5}{l}{\textbf{\textit{AGIEval}}}\\
AGIEval & 1 & Flex-math-2x7B-1T & 2048 & 0.4072 \\
AGIEval & 2 & Flex-news-2x7B-1T & 32 & 0.4048 \\
AGIEval & 3 & Flex-creative-2x7B-1T & 512 & 0.4020 \\
AGIEval & 4 & Flex-reddit-2x7B-1T & 512 & 0.3983 \\
AGIEval & 5 & Flex-pes2o-2x7B-1T & 1024 & 0.3949 \\
AGIEval & 6 & Flex-code-2x7B-1T & 2048 & 0.3903 \\
\midrule
\multicolumn{5}{l}{\textbf{\textit{BBH}}}\\
BBH & 1 & Flex-math-2x7B-1T & 2048 & 0.4289 \\
BBH & 2 & Flex-code-2x7B-1T & 2048 & 0.3776 \\
BBH & 3 & Flex-pes2o-2x7B-1T & 2048 & 0.3653 \\
BBH & 4 & Flex-news-2x7B-1T & 64 & 0.3646 \\
BBH & 5 & Flex-reddit-2x7B-1T & 2048 & 0.3613 \\
BBH & 6 & Flex-creative-2x7B-1T & 4 & 0.3578 \\
\midrule
\multicolumn{5}{l}{\textbf{\textit{MMLU}}}\\
MMLU & 1 & Flex-creative-2x7B-1T & 4 & 0.5605 \\
MMLU & 2 & Flex-news-2x7B-1T & 2 & 0.5596 \\
MMLU & 3 & Flex-math-2x7B-1T & 2048 & 0.5557 \\
MMLU & 4 & Flex-code-2x7B-1T & 4 & 0.5516 \\
MMLU & 5 & Flex-pes2o-2x7B-1T & 2 & 0.5513 \\
MMLU & 6 & Flex-reddit-2x7B-1T & 1024 & 0.5474 \\
\midrule
\multicolumn{5}{l}{\textbf{\textit{MMLU-Pro}}}\\
MMLU-Pro & 1 & Flex-news-2x7B-1T & 1 & 0.2636 \\
MMLU-Pro & 2 & Flex-math-2x7B-1T & 2048 & 0.2598 \\
MMLU-Pro & 3 & Flex-creative-2x7B-1T & 16 & 0.2595 \\
MMLU-Pro & 4 & Flex-pes2o-2x7B-1T & 1024 & 0.2523 \\
MMLU-Pro & 5 & Flex-code-2x7B-1T & 4 & 0.2498 \\
MMLU-Pro & 6 & Flex-reddit-2x7B-1T & 256 & 0.2457 \\
\bottomrule
\end{tabular*}
\caption{
Top-6 (all) unique experts per benchmark. For each expert, the best-performing LoRA rank (rank $\leq 2^{11}$) is selected, with ties broken by lowest rank.
For MC9, all unique experts achieve performance within 1.1 percentage points of the maximum score, despite requiring LoRA ranks that vary by over three orders of magnitude.
Expert mapping: news (News), code (Code), creative (Creative Writing), pes2o (Academic), reddit (Reddit) and math (Math)
}
\label{tab:topk_unique_experts}
\end{table*}

\section{Sensitivity Analysis \& Typical Peak Rank}
\label{sec:appendix-rank-senstivity}

\subsection{Rank Sensitivity Analysis}
\label{sec:appendix-rank-sensitivity-analysis}

To characterize how performance varies with LoRA rank, we analyze rank sensitivity using simple linear regression between evaluation score and $\log_2$ LoRA rank.
For each model family and evaluation group, we fit
$s(r) = \alpha + \beta \log_2 r$
where $s(r)$ denotes the observed score at rank $r$.
The coefficient $\beta$ provides a coarse summary of how performance changes with increasing rank: positive values indicate consistent gains, while values near zero or negative suggest diminishing returns.
This analysis is applied to both individual experts and combined mixture-of-experts models to enable comparison across settings.
Detailed results are provided in Tables \ref{tab:rank_sensitivity_summary}, \ref{tab:rank_sensitivity_detailed}, \ref{tab:expert_rank_sensitivity} and \ref{tab:expert_evalgroup_rank_sensitivity}.

\begin{table*}
\centering
\begin{tabular}{lrrrr}
\toprule
\textbf{Model Group} & \textbf{median} & \textbf{min} & \textbf{max} & $n$ \textbf{groups} \\
\midrule
\multicolumn{5}{l}{\textbf{\textit{FlexMoRE 7x7B-IT}}} \\
FlexMoRE-a2 & 0.0003 & -0.0042 & 0.0104 & 6 \\
FlexMoRE-a4 & 0.0030 & -0.0031 & 0.0074 & 6 \\
FlexMoRE-a7 & 0.0008 & -0.0083 & 0.0036 & 6 \\
\midrule
\multicolumn{5}{l}{\textbf{\textit{Individual Experts 2x7B-IT}}} \\
Flex-math & 0.0018 & -0.0010 & 0.0108 & 6 \\
Flex-pes2o & 0.0001 & -0.0003 & 0.0013 & 6 \\
Flex-code & -0.0000 & -0.0027 & 0.0030 & 6 \\
Flex-reddit & -0.0003 & -0.0065 & 0.0017 & 6 \\
Flex-creative & -0.0010 & -0.0018 & 0.0002 & 6 \\
Flex-news & -0.0016 & -0.0026 & 0.0001 & 6 \\
\midrule
\multicolumn{5}{l}{\textbf{\textit{Experts}}} \\
All experts & -0.0002 & -0.0018 & 0.0026 & 6 \\
\bottomrule
\end{tabular}
\caption{
Median rank sensitivity (slope per log$_2$ rank) across six benchmarks, computed via linear regression using at least four rank points per task (rank $2^{0}$ - $2^{14}$).
Positive values indicate that increasing LoRA rank consistently improves performance.
Each row summarizes the distribution of rank sensitivity across evaluation groups, where the median reflects the typical effect and the min/max capture task-dependent variability.
FlexMoRE models exhibit positive rank sensitivity, with the strongest effect at an intermediate number of active experts (a4), indicating diminishing returns at higher expert counts.
In contrast, expert-only models show near-zero or negative rank sensitivity, suggesting limited benefit from increased LoRA rank.
}
\label{tab:rank_sensitivity_summary}
\end{table*}

\begin{table*}
\centering
\begin{tabular}{llrrr}
\toprule
\textbf{Model Group} & \textbf{Benchmark} & \textbf{Slope per log$_2$ rank} & \textbf{Pearson r }& \textbf{n points} \\
\midrule
\multicolumn{5}{l}{\textbf{\textit{AGIEval}}} \\
Experts & AGIEval & 0.0001 & 0.0385 & 90 \\
FlexMoRE-a2 & AGIEval & 0.0006 & 0.3527 & 15 \\
FlexMoRE-a4 & AGIEval & 0.0026 & 0.9105 & 15 \\
FlexMoRE-a7 & AGIEval & 0.0004 & 0.2844 & 15 \\
\midrule
\multicolumn{5}{l}{\textbf{\textit{BBH}}} \\
Experts & BBH & 0.0026 & 0.4765 & 90 \\
FlexMoRE-a2 & BBH & 0.0104 & 0.9433 & 15 \\
FlexMoRE-a4 & BBH & 0.0074 & 0.9608 & 15 \\
FlexMoRE-a7 & BBH & 0.0036 & 0.9081 & 15 \\
\midrule
\multicolumn{5}{l}{\textbf{\textit{GEN5}}} \\
Experts & GEN5 & -0.0018 & -0.4110 & 90 \\
FlexMoRE-a2 & GEN5 & -0.0042 & -0.8594 & 15 \\
FlexMoRE-a4 & GEN5 & -0.0031 & -0.8410 & 15 \\
FlexMoRE-a7 & GEN5 & -0.0083 & -0.7460 & 15 \\
\midrule
\multicolumn{5}{l}{\textbf{\textit{MC9}}} \\
Experts & MC9 & -0.0010 & -0.3110 & 90 \\
FlexMoRE-a2 & MC9 & -0.0025 & -0.5318 & 15 \\
FlexMoRE-a4 & MC9 & 0.0029 & 0.8473 & 15 \\
FlexMoRE-a7 & MC9 & 0.0020 & 0.7162 & 15 \\
\midrule
\multicolumn{5}{l}{\textbf{\textit{MMLU}}} \\
Experts & MMLU & -0.0005 & -0.2409 & 90 \\
FlexMoRE-a2 & MMLU & -0.0001 & -0.0642 & 15 \\
FlexMoRE-a4 & MMLU & 0.0031 & 0.8983 & 15 \\
FlexMoRE-a7 & MMLU & 0.0004 & 0.2032 & 15 \\
\midrule
\multicolumn{5}{l}{\textbf{\textit{MMLU-Pro}}} \\
Experts & MMLU-Pro & 0.0000 & 0.0074 & 90 \\
FlexMoRE-a2 & MMLU-Pro & 0.0010 & 0.6395 & 15 \\
FlexMoRE-a4 & MMLU-Pro & 0.0031 & 0.8800 & 15 \\
FlexMoRE-a7 & MMLU-Pro & 0.0012 & 0.5407 & 15 \\
\bottomrule
\end{tabular}
\caption{
Per evaluation-group rank sensitivity (slope per log$_2$ rank), Pearson $r$, and number of evaluated ranks.
Results are computed for ranks $2^{0}$ - $2^{14}$.
Rank sensitivity varies substantially across evaluation groups and model families.
Reasoning-heavy benchmarks such as BBH exhibit strong and consistent positive rank sensitivity in FlexMoRE models ($r \approx 0.9$), indicating that increased LoRA rank provides effective additional capacity.
In contrast, expert-only models show weak or near-zero sensitivity across groups, suggesting that rank effects are largely noise-dominated without active-expert routing.
Several knowledge-oriented benchmarks (e.g., GEN5, MC9) exhibit negative rank sensitivity, consistent with diminishing returns or overfitting at higher ranks.
}
\label{tab:rank_sensitivity_detailed}
\end{table*}

\begin{table*}
\centering
\begin{tabular}{lrrr}
\toprule
\textbf{Expert} & \textbf{Slope per log$_2$ rank} & \textbf{Pearson r} & \textbf{n points} \\
\midrule
\multicolumn{4}{l}{\textbf{\textit{Code}}} \\
Flex-code-2x7B-1T & -0.0000 & -0.0240 & 15 \\
\multicolumn{4}{l}{\textbf{\textit{Creative Writing}}} \\
Flex-creative-2x7B-1T & -0.0009 & -0.8510 & 15 \\
\multicolumn{4}{l}{\textbf{\textit{Math}}} \\
Flex-math-2x7B-1T & 0.0028 & 0.9550 & 15 \\
\multicolumn{4}{l}{\textbf{\textit{News}}} \\
Flex-news-2x7B-1T & -0.0013 & -0.8590 & 15 \\
\multicolumn{4}{l}{\textbf{\textit{Academic}}} \\
Flex-pes2o-2x7B-1T & 0.0003 & 0.4480 & 15 \\
\multicolumn{4}{l}{\textbf{\textit{Reddit}}} \\
Flex-reddit-2x7B-1T & -0.0014 & -0.5660 & 15 \\
\bottomrule
\end{tabular}
\caption{
Per-expert rank sensitivity (slope per log$_2$ rank) computed by linear regression between LoRA rank ($2^0$--$2^{14}$) and the aggregated average evaluation score (Avg\_mean).
Across experts, rank sensitivity exhibits substantial heterogeneity (median $-0.0004$, 25th/75th percentiles [$-0.0012$, $0.0002$], minimum $-0.0014$, maximum $0.0028$).
Only the math-specialized expert shows a strong and consistent positive rank effect, while most experts exhibit weak or negative rank sensitivity.
}
\label{tab:expert_rank_sensitivity}
\end{table*}

\begin{table*}
\centering
\begin{tabular}{llrrr}
\toprule
\textbf{Expert} & \textbf{Benchmark} & \textbf{Slope per log$_2$ rank} & \textbf{Pearson r} & \textbf{n points} \\
\midrule
\multicolumn{5}{l}{\textbf{\textit{Code}}} \\
Flex-code-2x7B-1T & MC9 & 0.0000 & 0.0260 & 15 \\
Flex-code-2x7B-1T & GEN5 & -0.0027 & -0.8400 & 15 \\
Flex-code-2x7B-1T & AGIEval & 0.0007 & 0.5620 & 15 \\
Flex-code-2x7B-1T & BBH & 0.0030 & 0.9800 & 15 \\
Flex-code-2x7B-1T & MMLU & -0.0011 & -0.9140 & 15 \\
Flex-code-2x7B-1T & MMLU-Pro & -0.0001 & -0.1160 & 15 \\
\midrule
\multicolumn{5}{l}{\textbf{\textit{Creative Writing}}} \\
Flex-creative-2x7B-1T & MC9 & -0.0011 & -0.7580 & 15 \\
Flex-creative-2x7B-1T & GEN5 & -0.0004 & -0.7370 & 15 \\
Flex-creative-2x7B-1T & AGIEval & 0.0002 & 0.3230 & 15 \\
Flex-creative-2x7B-1T & BBH & -0.0011 & -0.8230 & 15 \\
Flex-creative-2x7B-1T & MMLU & -0.0018 & -0.9000 & 15 \\
Flex-creative-2x7B-1T & MMLU-Pro & -0.0010 & -0.8840 & 15 \\
\midrule
\multicolumn{5}{l}{\textbf{\textit{Math}}} \\
Flex-math-2x7B-1T & MC9 & 0.0010 & 0.8220 & 15 \\
Flex-math-2x7B-1T & GEN5 & -0.0010 & -0.6550 & 15 \\
Flex-math-2x7B-1T & AGIEval & 0.0024 & 0.9280 & 15 \\
Flex-math-2x7B-1T & BBH & 0.0108 & 0.9230 & 15 \\
Flex-math-2x7B-1T & MMLU & 0.0016 & 0.8920 & 15 \\
Flex-math-2x7B-1T & MMLU-Pro & 0.0021 & 0.9260 & 15 \\
\midrule
\multicolumn{5}{l}{\textbf{\textit{News}}} \\
Flex-news-2x7B-1T & MC9 & -0.0023 & -0.9260 & 15 \\
Flex-news-2x7B-1T & GEN5 & 0.0001 & 0.1730 & 15 \\
Flex-news-2x7B-1T & AGIEval & -0.0026 & -0.8600 & 15 \\
Flex-news-2x7B-1T & BBH & -0.0001 & -0.1040 & 15 \\
Flex-news-2x7B-1T & MMLU & -0.0017 & -0.8990 & 15 \\
Flex-news-2x7B-1T & MMLU-Pro & -0.0015 & -0.8960 & 15 \\
\midrule
\multicolumn{5}{l}{\textbf{\textit{Academic}}} \\
Flex-pes2o-2x7B-1T & MC9 & -0.0003 & -0.2790 & 15 \\
Flex-pes2o-2x7B-1T & GEN5 & -0.0000 & -0.0580 & 15 \\
Flex-pes2o-2x7B-1T & AGIEval & 0.0006 & 0.5030 & 15 \\
Flex-pes2o-2x7B-1T & BBH & 0.0013 & 0.8160 & 15 \\
Flex-pes2o-2x7B-1T & MMLU & 0.0000 & 0.0300 & 15 \\
Flex-pes2o-2x7B-1T & MMLU-Pro & 0.0002 & 0.2400 & 15 \\
\midrule
\multicolumn{5}{l}{\textbf{\textit{Reddit}}} \\
Flex-reddit-2x7B-1T & MC9 & -0.0033 & -0.6510 & 15 \\
Flex-reddit-2x7B-1T & GEN5 & -0.0065 & -0.7490 & 15 \\
Flex-reddit-2x7B-1T & AGIEval & -0.0008 & -0.3500 & 15 \\
Flex-reddit-2x7B-1T & BBH & 0.0017 & 0.8690 & 15 \\
Flex-reddit-2x7B-1T & MMLU & 0.0001 & 0.1190 & 15 \\
Flex-reddit-2x7B-1T & MMLU-Pro & 0.0003 & 0.2750 & 15 \\
\bottomrule
\end{tabular}
\caption{
Per-expert, per-evaluation-Benchmark rank sensitivity (slope per log$_2$ rank) computed via linear regression between LoRA rank ($2^0$–$2^{14}$) and each evaluation Benchmark’s mean score.
Slopes quantify the direction and magnitude of rank effects for each expert–task combination.
Reasoning-oriented benchmarks (e.g., BBH, AGIEval) exhibit consistently positive rank sensitivity for the Math expert, while knowledge-centric benchmarks (e.g., GEN5, MC9) often show weak or negative sensitivity across experts.
Across all expert–task pairs, the distribution of slopes spans from negative to strongly positive values, with the median near zero and a wide min–max range, highlighting strong expert–task interactions and non-uniform rank effects.
}
\label{tab:expert_evalgroup_rank_sensitivity}
\end{table*}

\subsection{Typical Peak Rank Per Task}
\label{sec:appendix-rank-sensitivity-analysis}

Because linear trends do not capture where performance saturates, we additionally report the rank at which peak performance is observed.
For each expert $e$ and evaluation group $g$, the peak rank is defined as
$r^*_{e,g} = \operatorname*{arg\,max}_{r \in \{2^0,\dots,2^{14}\}} s_{e,g}(r)$
computed directly from the observed scores without regression or smoothing.
In the case of ties, the lowest rank achieving the maximum score is selected.
We summarize peak-rank behavior using the median and interquartile range across experts. Results provided in Table \ref{tab:expert_typical_peak_log2_rank}.

\begin{table*}
\centering
\begin{tabular}{lrrrr}
\toprule
\textbf{Benchmark} & \textbf{Median} & \textbf{Q25} & \textbf{Q75} & $n$ \\
\midrule
MMLU & 2.00 & 1.25 & 8.00 & 6 \\
GEN5 & 5.00 & 4.25 & 5.75 & 6 \\
MMLU-Pro & 6.00 & 2.50 & 9.50 & 6 \\
MC9 & 6.50 & 3.75 & 7.00 & 6 \\
AGIEval & 9.50 & 9.00 & 11.50 & 6 \\
BBH & 11.50 & 7.25 & 12.00 & 6 \\
\midrule
\textbf{Avg.} & \textbf{9.00}  & \textbf{6.75} & \textbf{10.50} & 6 \\
\bottomrule
\end{tabular}
\caption{
Typical log$_2$ LoRA rank at which experts achieve peak performance.
For each expert $e$ and evaluation Benchmark $g$, the peak rank is defined as $r^*_{e,g}=\arg\max_{r\in\{2^0,\dots,2^{14}\}} s_{e,g}(r)$, computed directly from the observed scores after sorting by rank and resolving ties by selecting the lowest rank (no regression, smoothing, or normalization).
The table reports the median and interquartile range (25th–75th percentiles) of $\log_2 r^*_{e,g}$ across experts.
Peak performance typically occurs at moderate ranks: for the aggregated average (Avg), the median peak is at $\log_2 r=9$ with IQR $[6.75,10.50]$ (i.e., ranks $\approx 2^7$–$2^{10}$).
Knowledge-oriented benchmarks peak earlier (e.g., MMLU: median $\log_2 r=2$, IQR $[1.25,8.00]$; GEN5: median $\log_2 r=5$, IQR $[4.25,5.75]$),
while reasoning-heavy benchmarks peak at substantially higher ranks (e.g., BBH: median $\log_2 r=11.5$, IQR $[7.25,12.00]$).
}
\label{tab:expert_typical_peak_log2_rank}
\end{table*}

\end{document}